\documentclass[journal,twoside,web]{ieeecolor}

\usepackage{colortbl}

\usepackage{tmi}
\usepackage{cite}
\usepackage{amsmath,amssymb,amsfonts}
\usepackage{arydshln}
\usepackage{multirow}
\usepackage[noend]{algpseudocode}
\usepackage{algorithmicx,algorithm}
\usepackage{graphicx}
\usepackage{textcomp}
\usepackage{epstopdf}
\usepackage{booktabs}
\usepackage[section]{placeins}
\usepackage{stfloats}
\usepackage{bm}
\usepackage{xstring}
\usepackage{bbding}
\usepackage{color}
\usepackage{ulem}
\definecolor{darkcyan}{RGB}{0,138,218}
\usepackage[colorlinks=true, allcolors=darkcyan,urlcolor = black]{hyperref}
\usepackage[switch]{lineno}
\useunder{\uline}{\ul}{}

\def\BibTeX{{\rm B\kern-.05em{\sc i\kern-.025em b}\kern-.08em
	T\kern-.1667em\lower.7ex\hbox{E}\kern-.125emX}}

\begin{document}
\title{Cyclic Learning: Bridging Image-level Labels and Nuclei Instance Segmentation}
\author{Yang Zhou, Yongjian Wu, Zihua Wang, Bingzheng Wei, Maode Lai, Jianzhong Shou, Yubo Fan, and Yan Xu
	\thanks{This work is supported by the National Natural Science Foundation in China under Grant 62022010, the Beijing Natural Science Foundation-Haidian District Joint Fund in China under Grant L222032, the Beijing hope run special fund of cancer foundation of China under Grant LC2018L02, the Fundamental Research Funds for the Central Universities of China from the State Key Laboratory of Software Development Environment in Beihang University in China, the 111 Project in China under Grant B13003, the high performance computing (HPC) resources at Beihang University. \textit{(Yang Zhou, Yongjian Wu, and Zihua Wang contributed equally to this work. Corresponding author: Yan Xu.)} }
	\thanks{Y. Zhou, Y. Wu, Z. Wang, Y. Fan, and Y. Xu are with the School of Biological Science and Medical Engineering, State Key Laboratory of Software Development Environment, Key Laboratory of Biomechanics and Mechanobiology of Ministry of Education, Beijing Advanced Innovation Center for Biomedical Engineering, Beihang University, Beijing 100191, China (e-mail: ZhouYangBME@buaa.edu.cn; wuyongjian@buaa.edu.cn; wangzihua07@126.com; yubofan@buaa.edu.cn; xuyan04@gmail.com).}
	\thanks{B. Wei is with Xiaomi Corporation, Beijing 100085, China (e-mail: bingzhengwei@hotmail.com).}
	\thanks{Maode Lai is with Department of Pathology, School of Medicine, Zhejiang University, Zhejiang Provincial Key Laboratory of Disease Proteomics and Alibaba-Zhejiang University Joint Research Center of Future Digital Healthcare, Hangzhou 310053, China (e-mail: lmd@zju.edu.cn).}
    \thanks{Jianzhong Shou is with the Department of Urology, National Cancer Center, National Clinical Research Center for Cancer, Cancer Hospital, Chinese Academy of Medical Sciences and Peking Union Medical College, Chaoyang District, Beijing 100021, China (e-mail: shoujianzhong@cicams.ac.cn).}
 }
\maketitle
\begin{abstract}
Nuclei instance segmentation on histopathology images is of great clinical value for disease analysis. Generally, fully-supervised algorithms for this task require pixel-wise manual annotations, which is especially time-consuming and laborious for the high nuclei density. To alleviate the annotation burden, we seek to solve the problem through image-level weakly supervised learning, which is underexplored for nuclei instance segmentation. Compared with most existing methods using other weak annotations (scribble, point, etc.) for nuclei instance segmentation, our method is more labor-saving. The obstacle to using image-level annotations in nuclei instance segmentation is the lack of adequate location information, leading to severe nuclei omission or overlaps. In this paper, we propose a novel image-level weakly supervised method, called cyclic learning, to solve this problem. Cyclic learning comprises a front-end classification task and a back-end semi-supervised instance segmentation task to benefit from multi-task learning (MTL). We utilize a deep learning classifier with interpretability as the front-end to convert image-level labels to sets of high-confidence pseudo masks and establish a semi-supervised architecture as the back-end to conduct nuclei instance segmentation under the supervision of these pseudo masks. Most importantly, cyclic learning is designed to circularly share knowledge between the front-end classifier and the back-end semi-supervised part, which allows the whole system to fully extract the underlying information from image-level labels and converge to a better optimum. Experiments on three datasets demonstrate the good generality of our method, which outperforms other image-level weakly supervised methods for nuclei instance segmentation, and achieves comparable performance to fully-supervised methods.
\end{abstract}

\begin{IEEEkeywords}
Nuclei Segmentation, Weakly Supervised Learning, Image-level Annotation, Multi-task Learning, CNN Interpretability, Semi-supervised Learning
\end{IEEEkeywords}

\section{Introduction}
\label{sec:introduction}


\IEEEPARstart{H}{istopathology} images stained with hematoxylin and eosin (H\&E) provide primordial information for cancer diagnosis, prognosis, and treatment decisions \cite{srinidhi2021deep}. These images are commonly stained with hematoxylin and eosin (H\&E) for pathologist to ascertain important pathological feature. Nuclei segmentation is a crucial step in the automatic analysis of histopathology images, because the segmentation result reviews important features like average size, density, and nucleus-to-cytoplasm ratio that are closely related to the clinical diagnosis \cite{irshad2013methods}. Histopathologic nuclei instance segmentation belongs to dense instance segmentation. Compared with common instance segmentation in natural scenes, nuclei are close to each other spatially and even appear as clusters, making it hard to locate all of the target nuclei without omission. Current fully-supervised methods \cite{vuola2019mask,zhou2019cia} demand a great number of pixel-level masks to achieve promising performance. However, annotating accurate pixel-level masks for enormous dense nuclei is laborious and requires expert knowledge, which severely limits practical applications. In addition, this complex annotation task may introduce subjective bias, as different physicians may delineate it differently \cite{bilodeau2022microscopy}.

This obstacle motivates researchers to develop weakly supervised methods, such as point-annotated approaches \cite{zhao2020weakly,qu2020weakly}, scribble-annotated approaches \cite{lee2020scribble2label}, box-annotated approaches \cite{yang2018boxnet}, partially-annotated approaches \cite{qu2020weakly}, and image-level weakly supervised approaches \cite{bilodeau2022microscopy}. Compared with other annotations, image-level labels as weak supervision are much more labor-saving, reducing further the complexity and duration of the annotation task \cite{bilodeau2022microscopy}. Moreover, simpler annotation tasks reduce the inter-participant variability, thus training segmentation models with image-level labels reduces the subjective bias and the annotating errors \cite{bilodeau2022microscopy}. But to date, nuclei instance segmentation via image-level weakly supervised learning for histopathology images is underexplored. 

It is non-trivial to exploit image-level labels in a nuclei instance segmentation task. Existing image-level weakly supervised methods are proposed for natural instance segmentation and do not focus on dense instance circumstances like nuclei in histopathology images. These methods exploit neural networks (NNs) to convert coarse-grained image-level labels to location cues of instances \cite{zhou2018weakly,ge2019label}. Thereinto, NN interpreting methods are adopted as a bridge between the image-level label and pixel-wise instance segmentation. However, current image-level weakly supervised methods tend to ignore the fact that NN interpreting methods are only responsive to a hand of high-semantic instance regions used by the network to determine the image class, leading to severe omission when instances are dense nuclei. But still, the instances located by NN interpreting methods have a high probability of being true-positive. If we regard these locations with high-semantic information as a pseudo incomplete annotation of the entire nuclei family, it is intuitive to think of retrieving the omitted nuclei by semi-supervised learning \cite{zhou2018brief}. However, a rigorous exploration of this intuitive observation is severely lacking in the literature. An intractable hurdle is that it is difficult to train a well-performing semi-supervised model based solely on the knowledge passed by the pseudo incomplete annotation. Nevertheless, there is a potential solution to this problem: multi-task learning (MTL) \cite{zhang2021survey,vandenhende2021multi}. MTL is a learning paradigm to jointly learn shared information from multiple tasks to improve the performance of all tasks. It expands the knowledge sharing path through certain approaches and has been recently proved to be effective in semantic segmentation \cite{zhang2018joint} and instance segmentation \cite{pham2019jsis3d}. Inspired by these works, we adopt MTL to improve the knowledge sharing between the NN classification and the semi-supervised instance segmentation.

In this work, we present a flexible and general MTL method called cyclic learning to realize nuclei instance segmentation using only image-level labels. Cyclic learning splits the image-level nuclei instance segmentation task into two tasks: a front-end classification and a back-end semi-supervised instance segmentation. The front-end focuses on transforming the coarse-grained image-level annotations into pseudo masks of a portion of nuclei,  while the back-end concentrates on completing the instance predictions through semi-supervised learning. Most important of all, cyclic learning utilizes the advantage of MTL through circularly sharing knowledge between the front-end and the back-end task, facilitating the optimization of both. Eventually, cyclic learning improves the performance of the front-end and the back-end simultaneously. Theoretical analysis proves that circularly sharing knowledge ensures the whole image-level weakly supervised framework to converge to a better optimum. Extensive experiments indicate that the performance of cyclic learning is close to the fully-supervised method, indeed bridging image-level labels and nuclei instance segmentation. Code is released at \href{https://github.com/wuyongjianCODE/Cyclic}{https://github.com/wuyongjianCODE/Cyclic}.

The main contributions of this paper are summarized as follows.
\begin{itemize}
\item A novel weakly supervised method under image-level supervision is proposed for nuclei instance segmentation, which is less explored by any previous work, to further reduce the annotation effort.
\item We propose an MTL training method named cyclic learning to share knowledge circularly between the NN interpreting method and the semi-supervised method. This approach facilitates both methods and ultimately achieves image-level nuclei instance segmentation with performance comparable to fully-supervised counterparts. 
\item Extensive experiments demonstrate the generality of cyclic learning, which is adaptable to different feature extracting backbones, including both convolutional NN and transformer. Additionally, cyclic learning is compatible with different back-end segmentation architectures, such as Mask-RCNN, Hover-net, and KG Instance. 

\end{itemize}

\section{Related works}

\begin{figure*}[t]
\begin{center}
\centerline{\includegraphics[width=\textwidth]{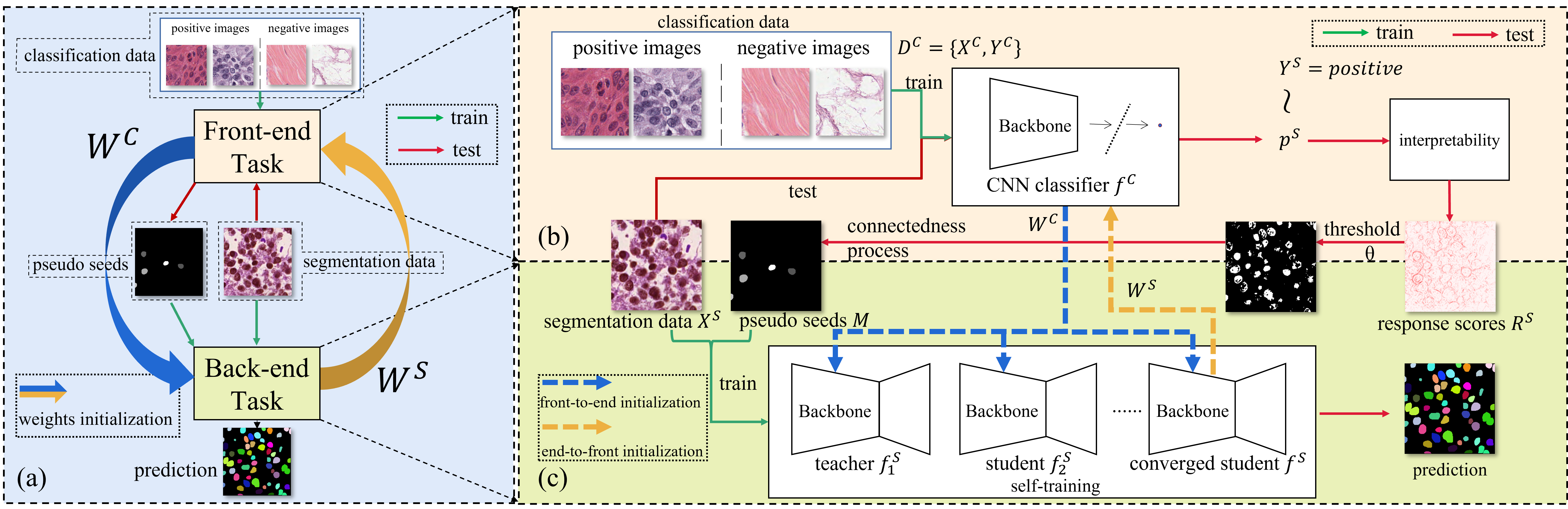}}
\caption{Illustration of the proposed method. \textcolor{darkcyan}{(a)} Cyclic learning, communicating tasks based on multi-task learning (MTL), circularly shares knowledge between the front-end task and the back-end task through weights until convergence. \textcolor{darkcyan}{(b)} The front-end trains a NN classifier $f^{C}$ to generate pseudo masks using a NN interpreting method, then passes the pseudo seeds and backbone weights to the back-end. \textcolor{darkcyan}{(c)} The back-end conducts self-training with the pseudo seeds, each student $f_{i}^{S}$ is initialized by the weights from $f^C$. The backbone weights of the best student $f^{S}$ are passed to $f^C$ to start a new cycle. The cycle stops when the whole system is converged.}
\label{fig3}
\end{center}
\end{figure*}

\subsection{Nuclei Instance Segmentation}\; Nuclei instance segmentation is a fundamental task in histopathology analysis and has attracted the attention of the research community for years. It belongs to dense instance segmentation. Traditional algorithms \cite{naik2008automated,veta2013automatic,jung2010unsupervised,zhou2015nuclei} for this problem have poor generalization ability, whereas recent deep learning-based methods are more robust and more effective \cite{chen2017dcan,vuola2019mask,zhou2019cia,xiao2021polar,graham2019hover,stringer2021cellpose,yi2019multi}. However, these fully-supervised methods require a large amount of training data, which are pixel-wisely annotated. Collecting such datasets requires expert knowledge, which is laborious and time-consuming. Some methods attempt to solve this problem by generating synthetic data as extra data for fully-supervised training \cite{kromp2021evaluation, mahmood2019deep}. However, most researchers turn to weakly-supervised learning for this task \cite{lee2020scribble2label, qu2020weakly}. WSPointA \cite{qu2019weakly}, WSMixedA \cite{qu2020nuclei}, and WNSeg \cite{liu2022weakly} are typical nuclei instance segmentation methods using weak point annotations for supervision. However, image-level annotation is not well exploited in this task because it lacks location information compared to the aforementioned annotations \cite{qu2020weakly}, causing instance omission and coarse boundaries in nuclei instance segmentation. Yang \textit{et al.} have tried to use image-level labels for nuclei segmentation and propose a domain adaptation framework \cite{yang2021minimizing}. However, their method requires a significant number of source pixel-level masks before it can leverage image-level labels for target adaptation, whereas our proposed method can be trained from scratch using only image-level labels. Bilodeau \textit{et al.} propose MICRA-Net \cite{bilodeau2022microscopy} for cell segmentation using image-level weak supervision across various microscopic imaging modalities, including stimulated emission depletion (STED), fluorescence, and phase contrast. But H\&E stained histopathology images are more complicated in the background than ordinary microscopic images. In this study, we manage to tackle the image-level weakly supervised nuclei instance segmentation with the proposed MTL method called cyclic learning.

\subsection{Weakly Supervised Segmentation}\; To address the scarcity of pixel-wise data, weakly supervised approaches have been proposed for semantic and instance segmentation. Some methods try to use inexact annotations, like image-level labels \cite{wei2018revisiting,yao2021non}, scribbles \cite{lin2016scribblesup,tang2018normalized}, and bounding boxes \cite{hsu2019weakly,arun2020weakly}. Thereinto, image-level labels are easiest to obtain and many related weakly supervised methods utilize convolutional neural networks (CNN) interpreting approaches \cite{zhou2018weakly,ahn2019weakly,ge2019label}. Some other methods seek to exploit incomplete annotations with semi-supervised frameworks \cite{li2018weakly,guerrero2019weakly} like self-training \cite{riloff2003learning,zhou2020deep}. For dense instance segmentation, points \cite{qu2020weakly}, scribbles \cite{lee2020scribble2label}, bounding boxes \cite{yang2018boxnet}, and image-level labels \cite{bilodeau2022microscopy} have been used as weak supervision. Currently, most image-level weakly supervised methods are proposed for semantic segmentation, like OAA \cite{jiang2019integral}, MCIS \cite{sun2020mining}, and DRS \cite{kim2021discriminative}. In the natural scenario, Ahn \textit{et al.} propose IRNet for instance segmentation only with image-level labels \cite{ahn2019weakly}. For histopathology images, Xu \textit{et al.} present CAMEL using multiple-instance-learning technique for semantic segmentation \cite{xu2019camel}. Gu \textit{et al.} propose an image-level semantic segmentation method with transformer \cite{vaswani2017attention} for glands \cite{gu2022histosegrest}. Similarly, SwinMIL is a multiple-instance-learning method that also pioneers in leveraging transformer for image-level semantic segmentation on histopathology images \cite{qian2022transformer}. However, image-level annotations have not been tamed for nuclei instance segmentation in histopathology images, where nuclei are even denser spatially. Image-level methods designed for natural instance segmentation \cite{zhou2018weakly,ge2019label} do not perform well for nuclei instance segmentation in histopathology images.

\subsection{Multi-Task Learning}\; Multi-task learning (MTL) is a learning paradigm to jointly learn shared information from multiple tasks to improve the performance of all tasks \cite{caruana1997multitask,zhang2021survey,vandenhende2021multi}. All the tasks in an MTL system are learned together. Meanwhile, they are communicated to each other through certain knowledge sharing approaches \cite{zhang2021survey}. Past research has shown that MTL is more sample efficient than single-task learning in various application domains, such as natural language processing \cite{wang2018glue}, speech recognition \cite{deng2013new}, bioinformatics \cite{widmer2010leveraging}, and computer vision \cite{lapin2014scalable}. Most importantly, MTL has been proved to be effective at solving pixel-level prediction tasks like semantic segmentation \cite{zhang2018joint} and instance segmentation \cite{pham2019jsis3d,wen2020joint}. Inspired by these works, we cast the nuclei instance segmentation task under image-level supervision as two tasks: image classification and semi-supervised instance segmentation. Thus, we propose an MTL method called cyclic learning to bridge the coarse-grained annotations and the fine-grained nuclei predictions. Cyclic learning implements MTL by sharing the network backbones between the classification and semi-supervised instance segmentation.

\subsection{Interpreting Methods for Neural Networks}\; There are many interpreting methods that extract image regions that directly contribute to the network output for an image-level label to explain NN representations \cite{zhang2018visual}. Simonyan \textit{et al.} first explain the CNN through computing gradients of the score of a given CNN unit w.r.t. the input image \cite{simonyan2013deep}. Zhou \textit{et al.} propose class activation mapping (CAM) for CNN interpretability \cite{zhou2016learning}, and it is extended to Grad-CAM by Selvaraju \textit{et al.} \cite{selvaraju2017grad}. layer-wise relevance propagation (LRP) computes back-propagation for modified gradient function and focuses more on objects' edges compared with CAM\cite{binder2016layer}. As for the transformer backbone, attention rollout is proposed for interpretability by Abnar \textit{et al.} \cite{abnar2020quantifying}. As for interpreting methods used in image-level weakly supervised instance segmentation, they are required to transfer the annotations into local responses to find location cues of instance, among which peak response map (PRM) is representative \cite{zhou2018weakly}. Our proposed algorithm also uses a deep learning interpreting method to locate those nuclei with high confidence and complements the missing ones with semi-supervised learning and MTL.

\subsection{Semi-supervised learning}\; Semi-supervised learning attempts to exploit unlabeled data in addition to partial labeled data to improve learning performance \cite{zhou2018brief}. A simple way to implement semi-supervised learning is to first train classifiers on labeled data, and then use the predictions of the resulting classifiers to generate additional labeled data. The classifiers can then be re-trained on this pseudo-labeled data in addition to the existing labeled data. Such methods are known as wrapper methods \cite{van2020survey}. Self-training \cite{dopido2013semisupervised}, co-training \cite{ning2021review}, and boosting methods \cite{valizadegan2008semi} are typical wrapper methods and can be conveniently applied to off-the-shelf fully-supervised network architectures. Semi-supervised learning has been used in nuclei instance segmentation under partially
point-annotation supervision \cite{qu2020weakly}. As for image-level nuclei segmentation in this paper, the pseudo masks generated by NN interpretability are incomplete, ignoring a bunch of nuclei. The semi-supervised learning has the potential to complement the omission. In this work, we employ semi-supervised learning as a back end to dig out the omitted nuclei.

\section{Approach}
\label{sec:approach}

Nuclei instance segmentation aims to predict a pixel-wise mask for every single nucleus. It is challenging for a NN to tackle nuclei instance segmentation under image-level annotations, whose pixel-level location information is extremely indistinctive. A possible solution lies in converting the coarse-grained labels to fine-grained pseudo masks by utilizing the interpretability of NN. However, the generated pseudo masks are only a tiny fraction of the original nuclei, which seems rightly suitable for semi-supervised methods to handle with. However, the rigid combination of interpreting image-level labels with a NN classifier and semi-supervised learning has a lot of room for improvement. In this section, we explore a better way, namely cyclic learning, to combine the merits of the two. Specifically, cyclic learning organically combines the two parts based on MTL, which aims to leverage useful information contained in different tasks to improve the performance of all tasks at hand \cite{zhang2021survey}.

Cyclic learning consists of two tasks: (1) the front-end classification task that trains a NN classifier to convert the coarse-grained image-level labels to fine-grained pseudo masks using a deep learning interpreting method; (2) the back-end instance segmentation task to dig out the omitted nuclei with the semi-supervised learning method. The two tasks share knowledge through MTL in a cycling way (\autoref{fig3}). Cyclic learning establishes the knowledge sharing bridge through cyclic weights sharing between the identical feature extracting backbones at both ends. The final predictions are made by the back-end segmentation network with the optimized backbones. Our method is enlightened by Expectation-Maximization algorithm (EM) \cite{moon1996expectation} and works well with both CNN and transformer backbones. Moreover, cyclic learning is compatible with various back-end segmentation architectures. In the following sections, we mainly describe our method using a CNN as the feature extracting backbone. However, our method is also applicable to the scenario where transformer is used as the backbone.


\subsection{Front-End}

Image-level labels are coarse-grained annotations and are far from the requirement of fine-grained pixel-level dense nuclei segmentation. We need to convert the high-level semantic information from image-level labels to the low-level spatial information at the very beginning. A CNN classifier trained with image-level labels can be utilized for the conversion through CNN interpreting methods \cite{zhou2016learning}. Based on this, we devise a coarse-to-fine module to handle the front-end classification task and convert image-level labels to pixel-level pseudo masks of a part of nuclei with high confidence. 

The detailed design of the coarse-to-fine module is shown in \autoref{fig3}. A CNN classifier $f^{C}$ is trained on a dataset $\mathcal{D}^C = \left\{\left(X_{i}^{C}, Y_{i}^{C}\right)\right\}_{i=1}^{n_{C}}$, where $n_{C}$ denotes the number of data for the front-end classification task. After training $f^{C}$ by the cross entropy loss, we can get the global probability $p_{i}^{C}$ from $X_{i}^{C}$: $p_{i}^{C} = f^{C}\left( X_{i}^{C} \right), \;\; p_{i}^{C} \simeq Y_{i}^{C}.$

Utilizing off-the-shelf methods, we can decompose the global $p_{i}^{C}$ down to local response scores $R_{i}^{C}$ for the dimensions of $X_{i}^{C}$ by back-propagation. The local response score $R_{i}^{C}$ indeed contains the location cues for responsive nuclei. In this paper, we employ the layer-wise response propagation (LRP) \cite{binder2016layer} for this process.

Let $\mathcal{D}^S = \left\{\left(X_{i}^{S},Y_{i}^{S}\right)\right\}_{i=1}^{n_{S}}$ denote the dataset for the actual nuclei instance segmentation. So long as $\mathcal{D}^S$ is identical with $\mathcal{D}^C$ in term of data distribution, we can extract the local response scores $R_{i}^{S}$ for $X_{i}^{S}$. 

\begin{figure}[h]
\begin{center}
\centerline{\includegraphics[width=\columnwidth]{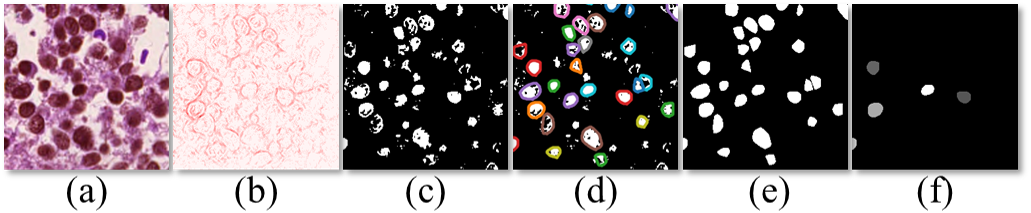}}
\caption{Pseudo masks generation. \textcolor{darkcyan}{(a)} Image. \textcolor{darkcyan}{(b)} $R$ generated by LRP. \textcolor{darkcyan}{(c)} Filtered $R$ with response threshold $\theta$. \textcolor{darkcyan}{(d)} Convex hull completion. \textcolor{darkcyan}{(e)} Morphological dilation and erosion. \textcolor{darkcyan}{(f)} Pseudo masks.}
\label{fig:seedgen}
\end{center}
\end{figure}

The primal $R_{i}^{S}$ is filled with noise. Thus, we set a response threshold $\theta$ to remove noise and explore connected components with high responses. The connected components are completed with convex hull \cite{graham1983finding} and morphological dilation and erosion \cite{gonzalez2009digital}. We randomly keep no more than $K$ remaining connected components as pseudo masks for each $X_{i}^{S}$. This process is illustrated in \autoref{fig:seedgen}.

\subsection{Back-End}

Existing image-level weakly supervised methods seldom pay attention to nuclei instance segmentation in histopathology images. The semantic information distributes unevenly in different nuclei, a large number of which are with low-semantic information. The location cues of these nuclei are inclined to be screened out by the noise filter. The pseudo masks generated by the front-end coarse-to-fine module also suffer from this defect. After being screened by $\theta$ and morphological process, the precision of the kept connect components is high but the recall could be rather low. Intuitively, regarding the generated pseudo masks as incomplete annotations, we may explore the missing nuclei by semi-supervised learning \cite{zhou2018brief}. 

Semi-supervised learning aims to exploit unlabeled data in addition to partially labeled data to improve the model performance without human intervention. Unlabeled data share the same cluster structure and lie on the same manifold in a geometric space as labeled ones, thus can be regarded as labeled data with perturbation. Training with them can build a more robust and representative model which yields more promising predictions \cite{zhou2018brief}.

We adopt a wrapper method, the self-training strategy, \cite{dopido2013semisupervised} for semi-supervised learning (see \autoref{fig3}). because the wrapper method can conveniently implement off-the-shelf fully-supervised network architectures. Let $\mathbf{M}_{i}=\left\{M_{1}, \ldots, M_{k_{i}}\right\}_{i}$, $i=1,2,...,n_{S}$, denote the pseudo masks generated by the front-end, where $k_{i}$ denotes the number of pseudo masks in $X_{i}^{S}$ and $k_{i}\leqslant K$. We call $\mathbf{M}_{i}$ the pseudo seeds. The self-training strategy can be summarized by the following steps. First, train a teacher model ${f}^{S}$ on $\left\{\left(X_{i}^{S}, \mathbf{M}_{i}\right)\right\}_{i=1}^{n_{s}}$. Without loss of generality, we employ Mask-RCNN \cite{he2017mask} as ${f}^{S}$. The modular design of Mask-RCNN allows for easy switching of the feature extracting backbone, which should be kept identical to that of the front-end network in our method to facilitate knowledge sharing (details in \ref{sec:cl}). The back-end segmentation architecture should have a feature extracting backbone compatible with the front-end classification. Any segmentation architecture with this characteristic can be used as ${f}^{S}$, including Hover-net \cite{graham2019hover} and KG Instance \cite{yi2019multi}. Denote the overall loss function of Mask-RCNN as $\mathcal{L}(\cdot)$ and the parameters of ${f}^{S}$ as $\mathbf{W}^S$, the objective is formularized as:
\begin{equation}
    {\mathbf{W}^S}' = \underset{\mathbf{W}^S}{\operatorname{argmin}} {\frac{1}{n_{s}} \sum_{i=1}^{n_{s}} \mathcal{L}\left(\mathbf{M}_{i}, f^{S}\left(X_{i}^{S} ; \mathbf{W}^{S}\right)\right)}.
\end{equation}
Secondly, use the convergent $f^{S}$ to predict a new pseudo mask set $\mathbf{M}'_{i}=\left\{M_{1},\ldots, M_{k'_{i}}\right\}_{i},i=1,2,..., n_{s}$.
Thirdly, based on the new pseudo masks set $\mathbf{M}'_{i}$ and the old one $\mathbf{M}_{i}$, learn an equal-or-better student ${f^{S}}'$ with a weighting parameter $\alpha$:
\begin{equation}
\begin{aligned}
    {\mathbf{W}^S}'' =& \underset{{\mathbf{W}^S}'}{\operatorname{argmin}} \frac{1}{n_{s}} \sum_{i=1}^{n_{s}} \bigg[ \alpha \mathcal{L}\left(\mathbf{M}_{i}, {f^{S}}'\left(X_{i}^{S}; {W^{S}}'\right)\right)\\&+(1-\alpha) \mathcal{L}\left(\mathbf{M}'_{i}, {f^{S}}'\left(X_{i}^{S}; {W^{S}}'\right)\right) \bigg].
\end{aligned}
\end{equation}
where $\alpha$ is a weighting parameter. Finally, extend $\mathbf{M}_{i}$ with $\mathbf{M}'_{i}$, i.e. $\mathbf{M}_{i} = \mathbf{M}_{i} \bigcup \mathbf{M}'_{i}$, let ${f^{S}}'$ be the new teacher and back to step 2 to generate new pseudo masks and train a new student. Repeat this process until the student is converged.

\subsection{Cyclic Learning}

\label{sec:cl}

Exploiting semi-supervised learning to make up for nuclei omission is intuitive. However, the rigid combination, i.e. Sharing knowledge only through pseudo seeds from the front-end classification, is superficial, which limits the performance of dense nuclei instance segmentation. To fully unleash the potential of image-level annotations, we propose a training method called cyclic learning (\autoref{fig3}).

Cyclic learning belongs to MTL techniques. Compared to the case where each task is solved separately, MTL methods have the potential for improved performance if the tasks at hand share complementary information \cite{zhang2021survey,vandenhende2021multi}. In the intuitive setting, the quality of generated pseudo seeds entirely relies on the front-end classification task, which is relatively independent of the back-end semi-supervised nuclei segmentation. The pseudo seeds are the only knowledge bridge between the two tasks. The front-end objective forces $f^C$ to learn semantic knowledge, whereas the back-end objective equips ${f}^{S}$ with location knowledge. 

Cyclic learning exploits the feature transformation approach \cite{zhang2021survey} from MTL. First, let the feature extractors in both $f^C$ and $f^S$ share the same backbone. After the convergence of the front-end $f^C$, the optimized parameters $\mathbf{W}_{1}^{C}$ of the feature extractor in $f^C$ are also used to initialize the back-end parameters besides generating pseudo seeds. Then, after self-training, the parameters of the convergent back-end extractor are in turn imposed on the front-end extractor to start a new round of classification training to yield a new set of pseudo seeds for the back-end. This bilateral knowledge sharing by weights forms a closed cycle and is circularly conducted until the whole system is fully optimized, yielding the final $f^S$ for dense nuclei instance segmentation with $\mathbf{W}_{m}^{S}$. In our setting, we adopt the prevailing ResNet \cite{he2016deep} as the default feature-extracting backbone without loss of generality. However, the newly proposed transformer \cite{vaswani2017attention} can also serve as the feature-extracting backbone in cyclic learning, as long as the backbones at both ends are consistent.

In the $i^{\text{th}}$ cycle, the semantic knowledge learned by the front-end $f^C$, which is implicit in $\mathbf{W}_{i}^{C}$, is used as an informative prior for the back-end. We could view the self-training process as a feature transformation $\Psi_{i}^{S}(\cdot)$. The backward semicycle can be formularized as:
\begin{equation}
    \mathbf{W}_{i}^{S}=\Psi_{i}^{S}\left(\mathbf{W}_{i}^{C}\right),
\end{equation}
where $i=1,2,...,m$. In return, the location knowledge complemented from the back-end turns into the prior for the front-end classification when $\mathbf{W}_{i}^{S}$ is transformed by the forward semicycle. Let the new round optimization for $f^C$ be $\Psi_{i}^{C}(\cdot)$, the forward semicycle can be formularized as:
\begin{equation}
    \mathbf{W}_{i+1}^{C}=\Psi_{i+1}^{C}\left(\mathbf{W}_{i}^{S}\right).
\end{equation}
Through the cyclic learning pair $\left(\Psi_{i}^{S}(\cdot),\Psi_{i+1}^{C}(\cdot)\right)$, $f^S$ can directly receive the semantic knowledge learned from the front-end classification task, while $f^C$ can be complemented and polished by the location knowledge provided by the back-end instance segmentation task. Repeating this pattern can further improve performance with the information circulated between two tasks.

Cyclic learning establishes a stronger information bridge besides the pseudo seeds for the cross-task talk between the front-end task and the back-end task, utilizing information from one task to refine the backbone of another task. In this way, the hidden information carried by image-level labels can be learned for mitigating the omission and nuclei overlaps. It should be noted that at the beginning of each cycle, the predictions from the former back-end are not extended to the set of pseudo seeds. This aims to manage the influence of introducing more pseudo seeds brought to the back-end semi-supervised module. Meanwhile, it verifies that cyclic learning indeed learns a better backbone through circularly sharing knowledge, which is expatiated in \autoref{sec:ablaiton}.

\section{Theoretical Analysis}
\label{sec:theoretical}

Casting the image-level nuclei instance segmentation task as a front-end task and a back-end task and training them iteratively can make the whole system converge to a better optimum. Inspired by Expectation-Maximization (EM) algorithm \cite{moon1996expectation,forbes2005convergence}, we give the theoretical proof here. 

Given the front-end data $\mathcal{D}^C = \left\{\left(X_{i}^{C}, Y_{i}^{C}\right)\right\}_{i=1}^{n_{C}}$ and the back-end data $\mathcal{D}^S = \left\{\left(X_{i}^{S}, Y_{i}^{S}\right)\right\}_{i=1}^{n_{S}}$ as the observed variables $\mathcal{D}$, the goal of our task is substantially to find $\mathbf{W}$, including $\mathbf{W}^{C}$ and $\mathbf{W}^{S}$, such that the joint probability density $\mathcal{P}(\mathcal{D} \mid \mathbf{W})$ is a maximum. Let $\mathbb{L}(\mathbf{W}\mid \mathcal{D})$ denote the maximum likelihood estimate of the backbone weights, the objective of cyclic learning can be essentially formularized as:
\begin{equation}
    \mathbf{W}_{best} = \underset{\mathbf{W}}{\operatorname{argmax}}\;\mathbb{L}(\mathbf{W}\mid \mathcal{D}).
\end{equation}
If we view the pixel-wise local response scores $\mathbf{R}^C = \left\{ R_{i}^{C} \right\} _{i=1}^{n_{C}}$ as the hidden variables,  we have:
\begin{equation}
\begin{aligned}
   \mathbb{L}(\mathbf{W}\mid\mathcal{D}^C) = & \log \mathcal{P}\left(\mathcal{D}^C\mid\mathbf{W}\right)\\ = & \log \sum_{\mathbf{R}^C} \mathcal{P} \left(\mathcal{D}^C,\mathbf{R}^C\mid\mathbf{W}\right).
\end{aligned}
\end{equation}
Thus, the objective of our task is indeed to find the maximum of $\log \sum_{\mathbf{R}^C} \mathcal{P} \left(\mathcal{D}^C,\mathbf{R}^C\mid\mathbf{W}\right)$. Assume that $\mathbf{R}^C$ match the distribution $Q(\mathbf{R}^C)$, which is initialized by $\mathbf{W}_{1}^{C}$ at the first cycle, we have:
\begin{equation}
\begin{aligned}
   \mathbb{L}(\mathbf{W}\mid\mathcal{D}^C) = & \log \sum_{\mathbf{R}^C} \mathcal{P} \left(\mathcal{D}^C,\mathbf{R}^C\mid\mathbf{W}\right)\\ = & \log \sum_{\mathbf{R}^C} \frac{Q(\mathbf{R}^C)\;\mathcal{P} \left(\mathcal{D}^C,\mathbf{R}^C\mid\mathbf{W}\right)}{Q(\mathbf{R}^C)}\\ = & \log \mathbb{E}_{Q(\mathbf{R}^C)}\left[ \frac{\mathcal{P} \left(\mathcal{D}^C,\mathbf{R}^C\mid\mathbf{W}\right)}{Q(\mathbf{R}^C)}\right] \\ \geqslant & \mathbb{E}_{Q(\mathbf{R}^C)} \left[ \log \frac{\mathcal{P} \left(\mathcal{D}^C,\mathbf{R}^C\mid\mathbf{W}\right)}{Q(\mathbf{R}^C)}\right],
\end{aligned}
\label{e-step}
\end{equation}
where $\mathbb{E}(\cdot)$ refers to the conditional expectation. The last relation of \eqref{e-step} used the Jensen's inequality \cite{kuczma1985introduction} and the equality holds only if $\frac{\mathcal{P} \left(\mathcal{D}^C,\mathbf{R}^C\mid\mathbf{W}\right)}{Q(\mathbf{R}^C)} = \mathbb{E}_{Q(\mathbf{R}^C)}\left[ \frac{\mathcal{P} \left(\mathcal{D}^C,\mathbf{R}^C\mid\mathbf{W}\right)}{Q(\mathbf{R}^C)}\right]=\mathcal{P}\left(\mathcal{D}^C\mid\mathbf{W}\right)=constant$, i.e. $Q(\mathbf{R}^C) = \frac{\mathcal{P} \left(\mathcal{D}^C,\mathbf{R}^C\mid\mathbf{W}\right)}{\mathcal{P}\left(\mathcal{D}^C\mid\mathbf{W}\right)} = \mathcal{P}\left(\mathbf{R}^C \mid \mathcal{D}^C,\mathbf{W}\right)$. Derivation \eqref{e-step} demonstrates that we can estimate the lower bound of $\mathbb{L}(\mathbf{W}\mid\mathcal{D}^C)$ by calculating $Q(\mathbf{R}^C)$ through $\mathcal{P}\left(\mathbf{R}^C \mid \mathcal{D}^C,\mathbf{W}\right)$. This process is essentially the optimization of the front-end classification aiming to yield better $Q(\mathbf{R}^C)$ through the weights from the last cycle. The last weights can be denoted as $\mathbf{W}_{i}$ and the new $Q(\mathbf{R}^C)$ can be denoted as $Q^{i+1}(\mathbf{R}^C)$ if the whole system is in the $(i+1)^{\text{th}}$ cycle. Derivation \eqref{e-step} is formally equivalent to the E-step in EM, which estimates the hidden variables given the observed data and current model parameters.

Since $\mathcal{D}^S$ is identical with $\mathcal{D}^C$ in term of data distribution, we can acquire better hidden variables $\mathbf{R}^S = \left\{ R_{i}^{S} \right\} _{i=1}^{n_{S}}$ through the front-end task, and use $Q(\mathbf{R}^S)$ to update $\mathbf{W}$ through maximizing the lower bound of $\mathbb{L}(\mathbf{W}_{i}\mid \mathcal{D}^S)$:
\begin{equation}
\begin{aligned}
  \mathbf{W}_{i+1} = & \underset{\mathbf{W}_{i}}{\operatorname{argmax}}\;\mathbb{L}(\mathbf{W}_{i}\mid \mathcal{D}^S) \\ = & \underset{\mathbf{W}_{i}}{\operatorname{argmax}}\; \mathbb{E}_{Q^{i+1}(\mathbf{R}^S)} \left[ \log \frac{\mathcal{P} \left(\mathcal{D}^S,\mathbf{R}^S\mid\mathbf{W}_{i}\right)}{Q^{i+1}(\mathbf{R}^S)}\right].
\end{aligned}
\label{m-step}
\end{equation}
In fact, \eqref{m-step} formularizes the process of learning the back-end nuclei instance segmentation, for that the set of pseudo seeds $\mathbf{M} = \left\{\mathbf{M}_{i}\right\}_{i=1}^{n^S}$ with high confidence is a projection of the hidden variables $\mathbf{R}^S$, including the information of $Q(\mathbf{R}^S)$. This process is formally equivalent to the M-step in EM, which maximizes the $\mathbb{L}(\mathbf{W}_{i}\mid \mathcal{D}^S)$ assuming the hidden variables are known. Through \eqref{e-step} and \eqref{m-step}, we have:
\begin{equation}
\begin{aligned}
    \mathbb{L}(\mathbf{W}_{i+1}\mid \mathcal{D}) = & \log \mathbb{E}_{Q^{i+1}(\mathbf{R})}\left[ \frac{\mathcal{P} \left(\mathcal{D},\mathbf{R}\mid\mathbf{W}_{i+1}\right)}{Q^{i+1}(\mathbf{R})}\right]\\ \geqslant & \mathbb{E}_{Q^{i+1}(\mathbf{R})} \left[ \log \frac{\mathcal{P} \left(\mathcal{D},\mathbf{R}\mid\mathbf{W}_{i+1}\right)}{Q^{i+1}(\mathbf{R})}\right]\\ \geqslant & \mathbb{E}_{Q^{i}(\mathbf{R})} \left[ \log \frac{\mathcal{P} \left(\mathcal{D},\mathbf{R}\mid\mathbf{W}_{i}\right)}{Q^{i}(\mathbf{R})}\right],
\end{aligned}
\label{iter-step}
\end{equation}
where the first inequality is a Jensen's inequality same as \eqref{e-step} and the second one is ensured by \eqref{m-step}. The inequality \eqref{iter-step} proves that introducing the hidden variables $\mathbf{R}$ ($\mathbf{R}^C$ and $\mathbf{R}^S$) and iteratively optimizing the front-end and the back-end ensures the increase of the maximum likelihood function. It states clearly that the lower bound of $\mathbb{L}(\mathbf{W}\mid\mathcal{D})$ in the $(i+1)^{\text{th}}$ cycle is greater than or equal to the one in the $i^{\text{th}}$ cycle, hence the system can converge to a better optimum.

From this point, we have proven that nuclei instance segmentation benefits from the combination of the front-end and the back-end. Furthermore, cyclic learning brings an extra advantage to this optimization process by circularly sharing knowledge between the front-end and the back-end.


\section{Experiments}

\begin{figure*}[t]
\begin{center}
\centerline{\includegraphics[width=\textwidth]{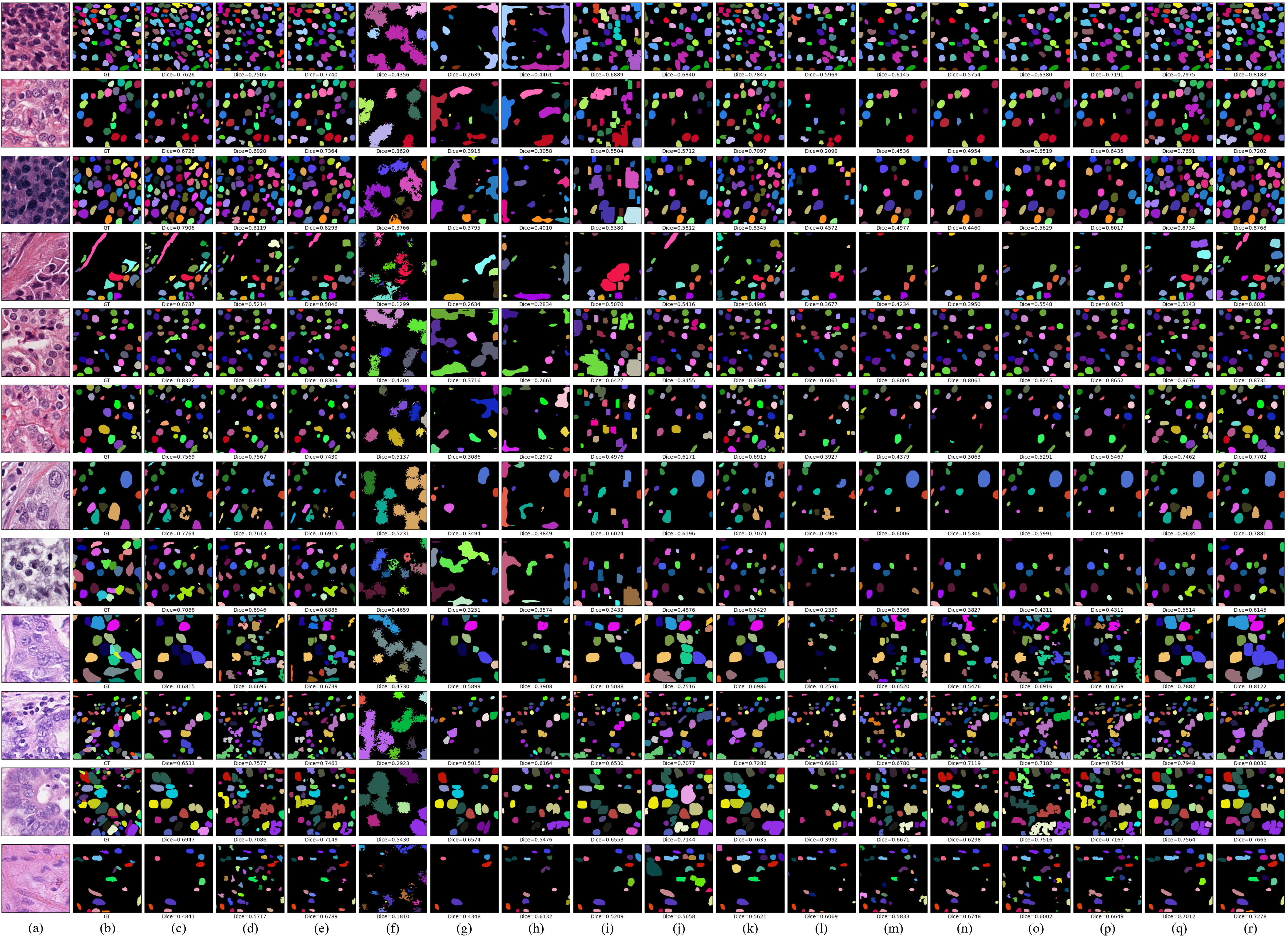}}
\caption{Comparison visualization. The $1^{\text{st}} \sim 4^{\text{th}}$ rows: MONu, the $5^{\text{th}} \sim 8^{\text{th}}$ row: CCRCC, the $9^{\text{th}} \sim 12^{\text{th}}$ row: CoNSeP. Column: \textcolor{darkcyan}{(a)} original images; \textcolor{darkcyan}{(b)} the ground truth; \textcolor{darkcyan}{(c)} WSPointA; \textcolor{darkcyan}{(d)} WSMixedA; \textcolor{darkcyan}{(e)} WNSeg; \textcolor{darkcyan}{(f)} PRM; \textcolor{darkcyan}{(g)} MDC-CAM; \textcolor{darkcyan}{(h)} MDC-Unet; \textcolor{darkcyan}{(i)} OAA; \textcolor{darkcyan}{(j)} CAMEL; \textcolor{darkcyan}{(k)} IRNet; \textcolor{darkcyan}{(l)} MCIS; \textcolor{darkcyan}{(m)} DRS; \textcolor{darkcyan}{(n)} NSROM; \textcolor{darkcyan}{(o)} SwinMIL; \textcolor{darkcyan}{(p)} MICRA-Net; \textcolor{darkcyan}{(q)} our cyclic learning; \textcolor{darkcyan}{(r)} fully-supervised Mask-RCNN. Different colors mark different instances. Color mapping is adopted between the ground truth and the prediction with false positive instances being randomly colored.}
\label{fig:comparison}
\end{center}
\end{figure*}

\subsection{Datasets}

\subsubsection{MONu}
MONu consists of 30 nuclei images of size $1000\times1000$ with 658 nuclei per image on average \cite{kumar2017dataset}. 
\textbf{Segmentation:} we use MONu images as the back-end dataset $\mathcal{D}^S$. Each of them is cropped from a region with dense nuclei of an original whole slide image (WSI) from TCGA \cite{tomczak2015cancer}. We follow Kumar \textit{et al.} \cite{kumar2017dataset} and use 16 images for training and the rest 14 for testing. Four images are randomly chosen for validation at each training epoch. We extract 16 overlapped image patches of size $256\times256$ from each image, and randomly crop them to $224\times224$ as the back-end input. It is noted that pixel-level ground truths are only used for evaluation, not for training.
\textbf{Classification:} for the front-end classification, two kinds of tiles with either a positive or a negative label are sampled from the original 30 WSIs to generate $\mathcal{D}^C$. The positive tiles contain target nuclei whereas the negative do not. These tiles were given by two expert pathologists and calibrated by a third expert. The total number of cropped tiles is 198, i.e. the total number of the used image-level labels. The training set consists of 113 positive and 59 negative tiles, while the test set comprises 17 positive and 9 negative tiles. Thus, only 172 image-level labels are used for training while other annotations are used for evaluation, much less than fully-supervised methods, which cost 658 instance-wise annotations per image on average. The average size of training tiles is $3090 \times 2636$ while the average size of the test tiles is $2087 \times 1460$. Tiles are cropped into patches of size $224\times224$ as the front-end input $\mathcal{D}^C$ with $n_{C}=42989$, 26846 of which are positive and 16143 are negative.

\subsubsection{CCRCC}
CCRCC is a dataset for Clear Cell Renal Cell Carcinoma grading and segmentation, consisting of 1000 H\&E stained images cropped from 149 TCGA WSIs \cite{gao2021nuclei}. For this dataset, we keep the same pre-processing procedure as MONu if not mentioned specifically.
\textbf{Segmentation:} we use the original CCRCC images for the back-end segmentation. There are 800, 100, and 200 images in training, validation, and test sets. We only use the ground-truth nuclei masks for evaluation.
\textbf{Classification:} 105 positive and 153 negative tiles are for training while 7 positive and 13 negative tiles are for testing. To wit, 278 image-level labels are used totally, of which 258 are for training. These tiles are cropped to $224\times224$ as $\mathcal{D}^C$ with $n_{C}=136139$, 114056 of which are positive and 22083 are negative.

\subsubsection{CoNSeP}
CoNSeP is composed of 41 H\&E stained images, each of size $1000\times1000$ and cropped from 16 colorectal adenocarcinoma WSIs, and includes many different tissue components and nuclei types \cite{graham2019hover}. Additionally, there are many significantly overlapping nuclei with indistinct boundaries and there exist various artifacts in CoNSeP\cite{graham2019hover}, which make it a rather complex dataset. For this dataset, we keep the same pre-processing procedure as MONu if not mentioned specifically.
\textbf{Segmentation:} we use the original dataset partition for the back-end segmentation: 26 images for training, 1 for validation, and the rest 14 for testing \cite{kumar2017dataset}. The ground-truth nuclei masks are only used for evaluation.
\textbf{Classification:} We use the same classification tiles as the MONu dataset to verify the practicability and the generality of our method under the circumstance of inconsistent data distributions between two ends.

\subsection{Evaluation Metrics and Implementation}

\subsubsection{Evaluation Metrics}\; We use four metrics to evaluate the performance of cyclic learning on nuclei instance segmentation, including Dice coefficient (Dice) \cite{dice1945measures}
Hausdorff distance (HD) \cite{huttenlocher1993comparing}
Aggregated Jaccard Index (AJI) \cite{kumar2017dataset}, and object-level Dice coefficient (Dice$_{obj}$) \cite{sirinukunwattana2015stochastic}. 

\subsubsection{Implementation Details}\; The model training is performed using four Nvidia RTX 2080 Ti GPUs, each with 11 GB of memory. We mainly use Resnet152 \cite{he2016deep} as the backbone of $f^C$ and $f^S$ and other comparison methods. At the first cycle, $f^C$ is initialized with ImageNet-pretrained weights. We train both $f^C$ and $f^S$ with SGD \cite{bottou2010large} setting the learning rate to $10^{-4}$ and $10^{-3}$ each and weight decay to $10^{-4}$. For the back-end Mask-RCNN, we set the maximum number of ground truth instances to 400 in one image, the non-max suppression threshold of RPN to 0.9 during training, and 0.7 during testing. As for other Mask-RCNN hyper-parameters, we keep the default setting used on COCO by He \textit{et al.} \cite{he2017mask}. The weighting parameter $\alpha$ is set to 0.66 after cross-validation. Other methods used for comparison follow the default setting described in their papers. All experiments are implemented on Tensorflow 1.9.0, Keras 1.19.0. After the convergence of the back-end self-training, we choose the model that has the best Dice value for the next cycle. We set the response threshold $\theta$ to 0.35 to filter the noise of LRP after cross-validation. The maximum of pseudo seeds $K$ is set as 20 per image if not specifically mentioned.

\begin{table*}[t]
\caption{Comparison with other methods on MONu, CCRCC, and CoNSeP. \#P is the abbreviation for the number of Parameters (auxiliary networks+the final segmentation network). The numerical numbers are mean values and the corresponding subscript is the standard deviation of cross-validation calculated on the test set. ``Fully'' and ``CL'' indicate the fully-supervised Mask-RCNN and cyclic learning, respectively.}
\label{tab:comparison}
\begin{center}
\begin{small}
\begin{sc}
\resizebox{\textwidth}{!}{
\begin{tabular}{cccccccccccccc}
\toprule[2pt]
\multicolumn{1}{c|}{\multirow{2}{*}{\textbf{Method}}} & \multicolumn{1}{c|}{\multirow{2}{*}{\textbf{\#P(M)}}} & \multicolumn{4}{c|}{\textbf{MONu}} & \multicolumn{4}{c|}{\textbf{ccRCC}} & \multicolumn{4}{c}{\textbf{Consep}} \\ \cline{3-14} 
\multicolumn{1}{c|}{} & \multicolumn{1}{c|}{} & \textbf{Dice↑} & \textbf{HD↓} & \textbf{AJI↑} & \multicolumn{1}{c|}{\textbf{Dice$_{obj}$↑}} & \textbf{Dice↑} & \textbf{HD↓} & \textbf{AJI↑} & \multicolumn{1}{c|}{\textbf{Dice$_{obj}$↑}} & \textbf{Dice↑} & \textbf{HD↓} & \textbf{AJI↑} & \textbf{Dice$_{obj}$↑} \\ \hline
\multicolumn{1}{c|}{\textbf{fully}} & \multicolumn{1}{c|}{0+191} & $0.7810_{0.0382}$ & $6.0010_{1.8405}$ & $0.6440_{0.0333}$ & \multicolumn{1}{c|}{$0.7073_{0.0430}$} & $0.8695_{0.0404}$ & $5.0124_{1.8414}$ & $0.7703_{0.0396}$ & \multicolumn{1}{c|}{$0.8857_{0.0372}$} & $0.8224_{0.0372}$ & $11.5738_{2.8249}$ & $0.5611_{0.0249}$ & $0.6923_{0.0313}$ \\ \hdashline
\multicolumn{14}{c}{\textbf{point-level}} \\
\multicolumn{1}{c|}{\textbf{WSPointA}\cite{qu2019weakly}} & \multicolumn{1}{c|}{126+141} & $0.6581_{0.0489}$ & $14.7131_{2.3438}$ & $0.4904_{0.0311}$ & \multicolumn{1}{c|}{$0.5063_{0.0475}$} & $0.6822_{0.0487}$ & $20.9699_{2.1645}$ & $0.4117_{0.0462}$ & \multicolumn{1}{c|}{$0.4664_{0.0435}$} & $0.6762_{0.0469}$ & $15.7728_{2.9504}$ & $0.4041_{0.0441}$ & $0.5691_{0.0495}$ \\
\multicolumn{1}{c|}{\textbf{WSMixedA}\cite{qu2020nuclei}} & \multicolumn{1}{c|}{145+136} & $0.7043_{0.0465}$ & $11.0351_{2.1261}$ & $0.4992_{0.0300}$ & \multicolumn{1}{c|}{$0.6297_{0.0530}$} & $0.7109_{0.0558}$ & $10.3355_{2.9818}$ & $0.4370_{0.0406}$ & \multicolumn{1}{c|}{$0.5590_{0.0437}$} & $0.6794_{0.0575}$ & $12.2930_{2.0174}$ & $0.4056_{0.0335}$ & $0.5528_{0.0521}$ \\
\multicolumn{1}{c|}{\textbf{WNSeg}\cite{liu2022weakly}} & \multicolumn{1}{c|}{156+125} & $0.7462_{0.0468}$ & $10.3211_{2.9866}$ & $0.5407_{0.0396}$ & \multicolumn{1}{c|}{$0.6310_{0.0504}$} & $0.7298_{0.0432}$ & $17.7965_{2.3111}$ & $0.4583_{0.0416}$ & \multicolumn{1}{c|}{$0.6223_{0.0549}$} & $0.7223_{0.0525}$ & $15.8248_{2.7782}$ & $0.4490_{0.0301}$ & $0.5565_{0.0403}$ \\ \hdashline
\multicolumn{14}{c}{\textbf{image-level}} \\
\multicolumn{1}{c|}{\textbf{PRM}\cite{zhou2018weakly}} & \multicolumn{1}{c|}{0+102} & $0.3774_{0.1094}$ & $13.1435_{7.3905}$ & $0.2368_{0.0921}$ & \multicolumn{1}{c|}{$0.3551_{0.1094}$} & $0.3826_{0.0981}$ & $15.2337_{7.6986}$ & $0.2486_{0.0953}$ & \multicolumn{1}{c|}{$0.3659_{0.0931}$} & $0.3421_{0.1028}$ & $20.1284_{6.0791}$ & $0.2213_{0.0803}$ & $0.3286_{0.0898}$ \\
\multicolumn{1}{c|}{\textbf{MDC-CAM}\cite{wei2018revisiting}} & \multicolumn{1}{c|}{0+117} & $0.2901_{0.0861}$ & $27.7113_{6.7394}$ & $0.1568_{0.0864}$ & \multicolumn{1}{c|}{$0.3247_{0.0862}$} & $0.3008_{0.0720}$ & $27.4172_{5.9619}$ & $0.1672_{0.0825}$ & \multicolumn{1}{c|}{$0.2812_{0.0855}$} & $0.3267_{0.0741}$ & $18.2859_{6.2630}$ & $0.1874_{0.0701}$ & $0.2810_{0.0726}$ \\
\multicolumn{1}{c|}{\textbf{MDC-Unet}\cite{wei2018revisiting}} & \multicolumn{1}{c|}{0+121} & $0.3376_{0.0718}$ & $23.2885_{6.4198}$ & $0.1881_{0.0756}$ & \multicolumn{1}{c|}{$0.3281_{0.0628}$} & $0.3178_{0.0787}$ & $21.7964_{5.4111}$ & $0.1698_{0.0619}$ & \multicolumn{1}{c|}{$0.2806_{0.0726}$} & $0.3415_{0.0825}$ & $18.0538_{5.4592}$ & $0.1892_{0.0677}$ & $0.3135_{0.0765}$ \\
\multicolumn{1}{c|}{\textbf{OAA}\cite{jiang2019integral}} & \multicolumn{1}{c|}{0+166} & $0.5245_{0.0601}$ & $11.6734_{5.3250}$ & $0.3623_{0.0512}$ & \multicolumn{1}{c|}{$0.5023_{0.0645}$} & $0.5528_{0.0639}$ & $12.1078_{5.2633}$ & $0.3321_{0.0535}$ & \multicolumn{1}{c|}{$0.5390_{0.0645}$} & $0.5229_{0.0671}$ & $14.8937_{5.9755}$ & $0.3418_{0.0554}$ & $0.5106_{0.0763}$ \\
\multicolumn{1}{c|}{\textbf{CAMEL}\cite{xu2019camel}} & \multicolumn{1}{c|}{58+133} & $0.6841_{0.0631}$ & $14.0346_{3.4252}$ & $0.5099_{0.0475}$ & \multicolumn{1}{c|}{$0.5387_{0.0728}$} & $0.7143_{0.0747}$ & $14.4721_{3.9532}$ & $0.4856_{0.0525}$ & \multicolumn{1}{c|}{$0.6129_{0.0719}$} & $0.6948_{0.0552}$ & $17.3845_{3.6672}$ & $0.4823_{0.0465}$ & $0.5559_{0.0580}$ \\
\multicolumn{1}{c|}{\textbf{IRNet}\cite{ahn2019weakly}} & \multicolumn{1}{c|}{92+205} & $0.6628_{0.0436}$ & $18.3001_{4.1924}$ & $0.4059_{0.0443}$ & \multicolumn{1}{c|}{$0.3693_{0.0539}$} & $0.6647_{0.0550}$ & $19.3727_{3.9120}$ & $0.3922_{0.0411}$ & \multicolumn{1}{c|}{$0.5491_{0.0538}$} & $0.6625_{0.0584}$ & $15.2935_{4.2217}$ & $0.4128_{0.0476}$ & $0.5014_{0.0545}$ \\
\multicolumn{1}{c|}{\textbf{MCIS}\cite{sun2020mining}} & \multicolumn{1}{c|}{110+168} & $0.4163_{0.0533}$ & $20.0253_{5.5066}$ & $0.2445_{0.0312}$ & \multicolumn{1}{c|}{$0.3566_{0.0402}$} & $0.3923_{0.0594}$ & $18.8812_{4.3318}$ & $0.2265_{0.0401}$ & \multicolumn{1}{c|}{$0.3716_{0.0461}$} & $0.4320_{0.0423}$ & $21.3852_{5.2914}$ & $0.2755_{0.0447}$ & $0.3640_{0.0561}$ \\
\multicolumn{1}{c|}{\textbf{DRS}\cite{kim2021discriminative}} & \multicolumn{1}{c|}{115+186} & $0.5832_{0.0671}$ & $11.3635_{4.2954}$ & $0.4011_{0.0489}$ & \multicolumn{1}{c|}{$0.5411_{0.0502}$} & $0.6028_{0.0641}$ & $10.2741_{3.0087}$ & $0.5121_{0.0484}$ & \multicolumn{1}{c|}{$0.5627_{0.0516}$} & $0.5975_{0.0611}$ & $14.4721_{3.9431}$ & $0.4260_{0.0436}$ & $0.5329_{0.0503}$ \\
\multicolumn{1}{c|}{\textbf{NSROM}\cite{yao2021non}} & \multicolumn{1}{c|}{92+167} & $0.5440_{0.0540}$ & $18.2956_{4.8248}$ & $0.3663_{0.0493}$ & \multicolumn{1}{c|}{$0.4833_{0.0564}$} & $0.5383_{0.0613}$ & $14.2331_{3.6919}$ & $0.4528_{0.0540}$ & \multicolumn{1}{c|}{$0.5129_{0.0585}$} & $0.5638_{0.0585}$ & $15.0444_{4.4203}$ & $0.3926_{0.0433}$ & $0.5025_{0.0574}$ \\
\multicolumn{1}{c|}{\textbf{SwinMIL}\cite{qian2022transformer}} & \multicolumn{1}{c|}{0+105} & $0.6510_{0.0588}$ & $20.4951_{3.0094}$ & $0.4826_{0.0323}$ & \multicolumn{1}{c|}{$0.4383_{0.0563}$} & $0.6762_{0.0585}$ & $15.7728_{2.6860}$ & $0.4041_{0.0308}$ & \multicolumn{1}{c|}{$0.5691_{0.0656}$} & $0.6726_{0.0595}$ & $20.5581_{3.7814}$ & $0.3984_{0.0490}$ & $0.5842_{0.0532}$ \\
\multicolumn{1}{c|}{\textbf{MICRA-Net}\cite{bilodeau2022microscopy}} & \multicolumn{1}{c|}{0+125} & $0.7127_{0.0628}$ & $16.8227_{2.0146}$ & $0.5593_{0.0447}$ & \multicolumn{1}{c|}{$0.5694_{0.0513}$} & $0.6890_{0.0681}$ & $15.3605_{2.8740}$ & $0.4305_{0.0417}$ & \multicolumn{1}{c|}{$0.5622_{0.0486}$} & $0.7244_{0.0413}$ & $15.1554_{2.8764}$ & $0.5886_{0.0303}$ & $0.5804_{0.0664}$ \\
\multicolumn{1}{c|}{\textbf{CL(Ours)}} & \multicolumn{1}{c|}{117+191} & $0.7739_{0.0336}$ & $5.6703_{1.3252}$ & $0.6357_{0.0339}$ & \multicolumn{1}{c|}{$0.7171_{0.0318}$} & $0.8669_{0.0327}$ & $3.7572_{1.5191}$ & $0.7664_{0.0238}$ & \multicolumn{1}{c|}{$0.8853_{0.0365}$} & $0.7846_{0.0301}$ & $15.2732_{1.9447}$ & $0.5192_{0.0278}$ & $0.6306_{0.0269}$ \\ \bottomrule[2pt]
\end{tabular}
}
\end{sc}
\end{small}
\end{center}
\end{table*}

\subsection{Comparison with Prior Methods}

We compare our method with 14 weakly supervised methods that mainly focus on panoptic or instance (cell, nuclei) segmentation: WSPointA (2019) \cite{qu2019weakly}, WSMixedA (2020) \cite{qu2020nuclei}, WNSeg (2022) \cite{liu2022weakly}, PRM (2018) \cite{zhou2018weakly}, MDC-CAM (2018) \cite{wei2018revisiting}, MDC-Unet (2018) \cite{wei2018revisiting}, OAA (2019) \cite{jiang2019integral}, CAMEL (2019) \cite{xu2019camel}, IRNet (2019) \cite{ahn2019weakly}, MCIS (2020) \cite{sun2020mining}, DRS (2021) \cite{kim2021discriminative}, NSROM (2021) \cite{yao2021non}, SwinMIL (2022) \cite{qian2022transformer}, and MICRA-Net (2022) \cite{bilodeau2022microscopy}. Among them, WSPointA, WSMixedA, and WNSeg use point annotations for supervision, while the others exploit image-level labels. We follow \cite{qu2019weakly} to generate the points annotation for three point-level methods by computing the central point of each nuclear mask. We alter the key hyper-parameters of these methods to fully unleash their potential on nuclei datasets. After cross-validation, the peak threshold in PRM is set to $0.5$; the saliency thresholds of MDC-CAM and MDC-Unet are set to 0.1 and 0.05, respectively; the attention threshold in OAA is set to 0.65; the attention threshold in MCIS is set to 0.3; the background cue weighting in DRS is set to 0.5; the attention threshold in NSROM is set to 0.5; the pretrained model used in SwinMIL is \textit{``swin\_tiny\_patch4\_window7\_224''}. We keep the default for other hyper-parameters or settings if not specifically mentioned.

\begin{table}[t]
\caption{Results of implementing cyclic learning and fully-supervised Mask-RCNN with different backbones.}
\label{tab:backbone}
\begin{center}
\begin{small}
\begin{sc}
\resizebox{\columnwidth}{!}{
\begin{tabular}{c|cccc}
\toprule[2pt]
\multirow{2}{*}{\textbf{Method}} & \multicolumn{4}{c}{\textbf{MONu}} \\ \cline{2-5} 
 & \textbf{Dice↑} & \textbf{HD↓} & \textbf{AJI↑} & \textbf{Dice$_{obj}$↑} \\ \hline
\textbf{fully-res50} & 0.7705 & 6.7604 & 0.6302 & 0.7040 \\
\textbf{CL-res50} & 0.7649 & 6.7387 & 0.6227 & 0.7008 \\ \hline
\textbf{fully-res101} & 0.7727 & 6.7860 & 0.6328 & 0.7093 \\
\textbf{CL-res101} & 0.7669 & 5.5467 & 0.6252 & 0.7196 \\ \hline
\textbf{fully-res152} & 0.7810 & 6.0010 & 0.6440 & 0.7073 \\
\textbf{CL-res152} & 0.7739 & 5.6703 & 0.6357 & 0.7171 \\ \bottomrule[2pt]
\end{tabular}
}
\end{sc}
\end{small}
\end{center}
\end{table}

The quantitative results are summarized in \autoref{tab:comparison}. \autoref{tab:comparison} also reports the numbers of parameters (auxiliary networks + the final segmentation network) of all methods, which are listed in column 2. It shows that our method (cyclic learning, CL) outperforms other image-level and even point-level weakly supervised methods by a large margin in nuclei instance segmentation with slender parameter growth. In fact, most of the image-level weakly supervised methods being compared pay less attention to dense nuclei instance segmentation. MICRA-Net focuses on image-level cell segmentation on microscopic imaging modalities like fluorescence and phase contrast\cite{bilodeau2022microscopy}. However, it falls short in terms of nuclei instance segmentation on H\&E stained images, which are known to have more complicated backgrounds. We are not aware of any existing method that focuses on this task. It is difficult to train a well-performing semi-supervised model based solely on the knowledge passed by the pseudo annotation from the classification task. Consequently, these methods underperform the fully-supervised one ($1^{st}$ row) a lot. By contrast, the three point-level methods ($2^{nd}$ to $4^{th}$ rows) generally achieve better results, but at the cost of higher annotation requirements. With a vast amount of easily-annotated image-level labels, cyclic learning even surpasses the point-level methods. The visualization results of the output samples from different methods are illustrated in \autoref{fig:comparison}. It could be observed that comparison methods fail to differentiate instances with different responses, leading to serious nuclei omission and overlaps, like PRM, MDC-CAM, MDC-Unet, and OAA. What's worse, most comparison methods are incapable of capturing accurate nuclei boundaries due to CNN interpreting noise. For instance, the boundaries predicted by PRM are severely rough. Contrarily, our method can locate more nuclei and segment relatively satisfactory boundaries. Most importantly, cyclic learning reaches a performance on par with the fully-supervised counterpart. We also implement our method and the fully-supervised Mask-RCNN with Resnet50 and Resnet101 on the MONu dataset. The results are listed in \autoref{tab:backbone}. It could be observed that cyclic learning is also effective when the backbone architecture alters, thus demonstrating the generality of our approach.

\subsection{Ablation Studies}
\label{sec:ablaiton}
We conduct ablation studies on the cyclic learning process, the front-end, and the back-end on the MONu dataset.

\begin{table}[t]
\caption{Detailed results of cycle learning in each cycle. C refers to ``Cycle''. The best results of each cycle are underlined and the best of all are marked in bold.}
\label{tab:cycle-student}
\begin{center}
\begin{small}
\begin{sc}
\resizebox{\columnwidth}{!}{%
\begin{tabular}{c|ccccc|ccccc}
\toprule[2pt]
\multirow{2}{*}{\textbf{C}} & \multicolumn{5}{c|}{\textbf{Dice↑}} & \multicolumn{5}{c}{\textbf{HD↓}} \\ \cline{2-11} 
 & $f_{1}^{S}$ & $f_{2}^{S}$ & $f_{3}^{S}$ & $f_{4}^{S}$ & $f_{5}^{S}$ & $f_{1}^{S}$ & $f_{2}^{S}$ & $f_{3}^{S}$ & $f_{4}^{S}$ & $f_{5}^{S}$ \\ \hline
\textbf{1} & 0.6616 & 0.7039 & {\ul 0.7181} & 0.6271 & 0.6494 & 9.0731 & {\ul 7.6412} & 8.0950 & 10.3806 & 9.6821 \\
\textbf{2} & 0.6793 & 0.7391 & {\ul 0.7484} & 0.7226 & 0.7322 & 8.0002 & 7.5747 & {\ul 6.8981} & 7.4078 & 7.5407 \\
\textbf{3} & 0.7046 & 0.7327 & 0.6803 & {\ul 0.7369} & 0.6781 & 9.1004 & 6.9650 & 7.9027 & {\ul 6.6245} & 9.0497 \\
\textbf{4} & 0.6908 & 0.7420 & 0.7339 & {\ul \textbf{0.7739}} & 0.7067 & 7.8425 & 6.9557 & 7.1788 & {\ul \textbf{5.6703}} & 8.2649 \\
\textbf{5} & 0.6868 & 0.7489 & 0.7576 & {\ul 0.7695} & 0.7505 & 8.0977 & 6.9250 & 6.7028 & {\ul 6.3828} & 6.6290 \\ \hline
\multirow{2}{*}{\textbf{C}} & \multicolumn{5}{c|}{\textbf{AJI↑}} & \multicolumn{5}{c}{\textbf{Dice$_{obj}$↑}} \\ \cline{2-11} 
 & $f_{1}^{S}$ & $f_{2}^{S}$ & $f_{3}^{S}$ & $f_{4}^{S}$ & $f_{5}^{S}$ & $f_{1}^{S}$ & $f_{2}^{S}$ & $f_{3}^{S}$ & $f_{4}^{S}$ & $f_{5}^{S}$ \\ \hline
\textbf{1} & 0.4991 & 0.5486 & {\ul 0.5655} & 0.4647 & 0.4884 & 0.5680 & 0.6368 & {\ul 0.6545} & 0.6143 & 0.6256 \\
\textbf{2} & 0.5208 & 0.5906 & {\ul 0.6030} & 0.5720 & 0.5827 & 0.6201 & 0.6797 & {\ul 0.6886} & 0.6740 & 0.6730 \\
\textbf{3} & 0.5479 & 0.5831 & 0.5228 & {\ul 0.5897} & 0.5251 & 0.6239 & 0.6569 & 0.6404 & {\ul 0.6945} & 0.6268 \\
\textbf{4} & 0.5324 & 0.5938 & 0.5876 & {\ul \textbf{0.6357}} & 0.5536 & 0.6026 & 0.6782 & 0.6779 & {\ul \textbf{0.7171}} & 0.6763 \\
\textbf{5} & 0.5280 & 0.6030 & 0.6144 & {\ul 0.6303} & 0.6061 & 0.5943 & 0.6833 & 0.6944 & {\ul 0.7078} & 0.6947 \\ \bottomrule[2pt]
\end{tabular}
}
\end{sc}
\end{small}
\end{center}
\end{table}

\subsubsection{Cyclic Learning} To investigate the effectiveness of cyclic learning, we evaluate the performance of the back-end self-training framework of each cycle, including the first teacher ($f_{1}^{S}$) to the convergent even overfit student ($f_{5}^{S}$). The numerical results are listed in \autoref{tab:cycle-student}. We can see that the best model of the back-end self-training can only reach 0.7181, 8.0950, 0.5655, and 0.6545 in Dice, HD, AJI, and Dice$_{obj}$ respectively at the initial cycle, where the pseudo seeds and only a kind of front-to-back knowledge carried by the convergent weights of the front-end are passed to the back-end. But after passing knowledge through cyclic learning, the performance of nuclei instance segmentation improves gradually and reaches convergence at the $4^{\text{th}}$ cycle, where the best model is compatible with the fully-supervised method.
\begin{table}[t]
\caption{Results of using different knowledge sharing modes. KSM, UNI, BIL, and C refer to ``knowledge sharing mode'', ``unilateral'', ``bilateral'', and ``Cycle'' respectively. The best results of unilateral knowledge sharing are underlined and the best of all are marked in bold.}
\label{tab:uni bi}
\begin{center}
\begin{small}
\begin{sc}
\resizebox{\columnwidth}{!}{
\begin{tabular}{c|c|c|cccc}
\toprule[2pt]
\multirow{2}{*}{\textbf{KSM}} & \multirow{2}{*}{\textbf{C}} & \textbf{front-end} & \multicolumn{4}{c}{\textbf{back-end}} \\ \cline{3-7} 
 &  & \textbf{F1-SCORE↑} & \textbf{Dice↑} & \textbf{HD↓} & \textbf{AJI↑} & \textbf{Dice$_{obj}$↑} \\ \hline
\multirow{5}{*}{\textbf{uni}} & \textbf{1} & 0.9294 & 0.7174 & 7.1655 & 0.5635 & 0.6571 \\
 & \textbf{2} & 0.9298 & 0.7416 & {\ul6.2628} & 0.5926 & 0.6414 \\
 & \textbf{3} & {\ul 0.9401} & {\ul0.7573} & 6.3181 & {\ul0.6124} & 0.6892 \\
 & \textbf{4} & 0.9354 & 0.7288 & 6.5722 & 0.5779 & {\ul0.6936} \\
 & \textbf{5} & 0.9344 & 0.7465 & 6.8385 & 0.5996 & 0.6682 \\ \hline
\multirow{5}{*}{\textbf{bil}} & \textbf{1} & 0.9250 & 0.7181 & 8.0950 & 0.5655 & 0.6545 \\
 & \textbf{2} & 0.9443 & 0.7484 & 6.8981 & 0.6030 & 0.6886 \\
 & \textbf{3} & 0.9414 & 0.7369 & 6.6245 & 0.5897 & 0.6945 \\
 & \textbf{4} & {\textbf{0.9540}} & {\textbf{0.7739}} & {\textbf{5.6703}} & {\textbf{0.6357}} & {\textbf{0.7171}} \\
 & \textbf{5} & 0.9321 & 0.7695 & 6.3828 & 0.6303 & 0.7078 \\ \bottomrule[2pt]
\end{tabular}
}
\end{sc}
\end{small}
\end{center}
\end{table}

\begin{table}[t]
\caption{The results of selecting random 20 ground-truth masks as ``pseudo'' seeds or 20 LRP-generated pseudo seeds for the back-end semi-supervised instance segmentation without cyclic learning.}
\label{tab:front-end}
\begin{center}
\begin{small}
\begin{sc}
\resizebox{0.8\columnwidth}{!}{
\begin{tabular}{c|c|c|c|c}
\toprule[2pt]
\textbf{Label} & \textbf{Dice↑} & \textbf{HD↓} & \textbf{AJI↑} & \textbf{Dice$_{obj}$↑} \\ \hline
\textbf{GT ×20} & 0.7176 & 7.0639 & 0.5636 & 0.6478 \\
\textbf{LRP ×20} & 0.7147 & 6.9826 & 0.5598 & 0.6354 \\ \bottomrule[2pt]
\end{tabular}
}
\end{sc}
\end{small}
\end{center}
\end{table}

\begin{table}[t]
\caption{Effectiveness of the back-end semi-supervised learning. The baseline is adopting a single Mask-RCNN as the back-end and its performance is listed in the table headers (B). ``+CL'' and ``+SSL,CL'' mean baseline adding cyclic learning and semi-supervised learning gradually. The best results of the comparison experiments are underlined and the best of all are marked in bold.}
\label{tab:back-end}
\begin{center}
\begin{small}
\begin{sc}
\resizebox{\columnwidth}{!}{
\begin{tabular}{c|cc|cc|cc|cc}
\toprule[2pt]
\multirow{2}{*}{\textbf{C}} & \multicolumn{2}{c|}{\textbf{Dice↑}(B:0.6616)} & \multicolumn{2}{c|}{\textbf{HD↓}(B:9.0731)} & \multicolumn{2}{c|}{\textbf{AJI↑}(B:0.4991)} & \multicolumn{2}{c}{\textbf{Dice$_{obj}$↑}(B:0.5680)} \\ \cline{2-9} 
 & \textbf{+CL} & \textbf{+SSL,CL} & \textbf{+CL} & \textbf{+SSL,CL} & \textbf{+CL} & \textbf{+SSL,CL} & \textbf{+CL} & \textbf{+SSL,CL} \\ \hline
\textbf{1} & 0.6616 & 0.7181 & 9.0731 & 8.0950 & 0.4991 & 0.5655 & 0.5680 & 0.6545 \\
\textbf{2} & 0.6781 & 0.7484 & 8.7326 & 6.8981 & 0.5167 & 0.6030 & 0.4923 & 0.6886 \\
\textbf{3} & 0.6660 & 0.7369 & 8.4793 & 6.6245 & 0.5047 & 0.5897 & 0.5773 & 0.6945 \\
\textbf{4} & {\ul 0.7194} & {\ul \textbf{0.7739}} & 7.8337 & {\ul \textbf{5.6703}} & {\ul 0.5655} & {\ul \textbf{0.6357}} & {\ul 0.6560} & {\ul \textbf{0.7171}} \\
\textbf{5} & 0.6866 & 0.7695 & {\ul 7.0114} & 6.3828 & 0.5279 & 0.6303 & 0.5513 & 0.7078 \\ \bottomrule[2pt]
\end{tabular}
}
\end{sc}
\end{small}
\end{center}
\end{table}

\begin{table*}[t]
\caption{Results of using different architectures as the back-end $f^S$ for cyclic learning (CL) and the ones of corresponding fully-supervised (FULLY) methods. The $1^{st}$ row and the $3^{rd}$ row are directly adopted from \cite{graham2019hover}. The $2^{nd}$ and the $4^{th}$ rows (REIMP.) present the results of our reimplementation. The best results of each architecture are marked in bold.}
\label{tab:diff fs}
\begin{center}
\begin{small}
\begin{sc}
\resizebox{\textwidth}{!}{
\begin{tabular}{c|cccc|cccc|cccc}
\toprule[2pt]
\multirow{2}{*}{\textbf{Method}} & \multicolumn{4}{c|}{\textbf{MONu}} & \multicolumn{4}{c|}{\textbf{ccRCC}} & \multicolumn{4}{c}{\textbf{Consep}} \\ \cline{2-13} 
 & \textbf{Dice↑} & \textbf{HD↓} & \textbf{AJI↑} & \textbf{Dice$_{obj}$↑} & \textbf{Dice↑} & \textbf{HD↓} & \textbf{AJI↑} & \textbf{Dice$_{obj}$↑} & \textbf{Dice↑} & \textbf{HD↓} & \textbf{AJI↑} & \textbf{Dice$_{obj}$↑} \\ \hline
\textbf{fully-mask-rcnn}\cite{he2017mask} & 0.7600 & \textbackslash{} & 0.5460 & \textbackslash{} & \textbackslash{} & \textbackslash{} & \textbackslash{} & \textbackslash{} & 0.7400 & \textbackslash{} & 0.4740 & \textbackslash{} \\
\textbf{fully-mask-rcnn}(reimp.) & \textbf{0.7810} & 6.0010 & \textbf{0.6440} & 0.7073 &\textbf{ 0.8695} & 5.0124 & \textbf{0.7703} & \textbf{0.8857} & \textbf{0.8224} & \textbf{11.5738} & \textbf{0.5611} & \textbf{0.6923} \\
\textbf{CL-mask-rcnn} & 0.7739 & \textbf{5.6703} & 0.6357 & \textbf{0.7171} & 0.8669 & \textbf{3.7572} & 0.7664 & 0.8853 & 0.7846 & 15.2732 & 0.5192 & 0.6306 \\ \hline
\textbf{fully-Hover-net}\cite{graham2019hover} & 0.8260 & \textbackslash{} & 0.6180 & \textbackslash{} & \textbackslash{} & \textbackslash{} & \textbackslash{} & \textbackslash{} & 0.8530 & \textbackslash{} & 0.5710 & \textbackslash{} \\
\textbf{fully-Hover-net}(reimp.) & \textbf{0.8215} & \textbf{3.1622} & 0.6044 & \textbf{0.7207} & \textbf{0.8707} & \textbf{4.0175} & 0.751 & \textbf{0.8639} & \textbf{0.8416} & \textbf{5.9078} & \textbf{0.6566} & \textbf{0.7235} \\
\textbf{CL-Hover-net} & 0.7922 & 4.2426 & \textbf{0.6059} & 0.7174 & 0.8675 & 4.2567 & \textbf{0.7541} & 0.8627 & 0.8274 & 6.5881 & 0.5621 & 0.7119 \\ \hline
\textbf{fully-KG Instance}\cite{yi2019multi} & \textbf{0.8117} & \textbf{7.8378} & \textbf{0.6831} & 0.7292 & \textbf{0.8701} & \textbf{3.8284} & \textbf{0.7743} & \textbf{0.8729} & \textbf{0.8367} & 9.2195 & \textbf{0.7193} & \textbf{0.7524} \\
\textbf{CL-KG Instance} & 0.7971 & 8.1927 & 0.6765 & \textbf{0.7296} & 0.8672 & 4.6758 & 0.7658 & 0.8718 & 0.8111 & \textbf{6.2361} & 0.6823 & 0.7229 \\ \bottomrule[2pt]
\end{tabular}
}
\end{sc}
\end{small}
\end{center}
\end{table*}

\begin{table*}[t]
\caption{Results of adopting swin-transformer \cite{liu2021swin} as the backbone of Mask-RCNN. ``Fully'' and ``CL'' indicate the fully-supervised Mask-RCNN and cyclic learning, respectively. The best results are marked in bold.}
\label{tab:transformer}
\begin{center}
\begin{small}
\begin{sc}
\resizebox{\textwidth}{!}{
\begin{tabular}{c|cccc|cccc|cccc}
\toprule[2pt]
\multirow{2}{*}{\textbf{Swin-Transformer}} & \multicolumn{4}{c|}{\textbf{MONu}} & \multicolumn{4}{c|}{\textbf{ccRCC}} & \multicolumn{4}{c}{\textbf{Consep}} \\ \cline{2-13} 
 & \textbf{Dice↑} & \textbf{HD↓} & \textbf{AJI↑} & \textbf{Dice$_{obj}$↑} & \textbf{Dice↑} & \textbf{HD↓} & \textbf{AJI↑} & \textbf{Dice$_{obj}$↑} & \textbf{Dice↑} & \textbf{HD↓} & \textbf{AJI↑} & \textbf{Dice$_{obj}$↑} \\ \hline
\textbf{fully} & \textbf{0.8061} & \textbf{3.1622} & \textbf{0.6159} & \textbf{0.7148} & \textbf{0.8296} & \textbf{4.2426} & \textbf{0.7143} & \textbf{0.7839} & \textbf{0.8047} & \textbf{6.6276} & \textbf{0.5732} & \textbf{0.6632} \\
\textbf{CL} & 0.7815 & 7.3796 & 0.5811 & 0.7019 & 0.8112 & 5.2956 & 0.7077 & 0.7522 & 0.7929 & 7.3071 & 0.5209 & 0.6303 \\ \bottomrule[2pt]
\end{tabular}
}
\end{sc}
\end{small}
\end{center}
\end{table*}

\begin{table}[t]
\caption{Results of adding synthetic data generated by DAT to the training set of MONu. The best results under the respective forms of supervision are marked in bold.}
\label{tab:synthetic data}
\begin{center}
\begin{small}
\begin{sc}
\resizebox{\columnwidth}{!}{
\begin{tabular}{c|c|cccc}
\toprule[2pt]
\multirow{2}{*}{\textbf{Method}} & \multirow{2}{*}{\textbf{DAT}\cite{mahmood2019deep}} & \multicolumn{4}{c}{\textbf{MONu}} \\ \cline{3-6} 
 &  & \textbf{Dice↑} & \textbf{HD↓} & \textbf{AJI↑} & \textbf{Dice$_{obj}$↑} \\ \hline
\multirow{2}{*}{\textbf{fully-maskrcnn}} &  & 0.7810 & 6.0010 & 0.6440 & 0.7073 \\
 & \checkmark & \textbf{0.788} & \textbf{5.624} & \textbf{0.6502} & \textbf{0.7254} \\ \hline
\multirow{2}{*}{\textbf{CL-maskrcnn}} &  & 0.7739 & 5.6703 & 0.6357 & 0.7171 \\
 & \checkmark & \textbf{0.7763} & \textbf{4.4721} & \textbf{0.6364} & \textbf{0.7183} \\ \bottomrule[2pt]
\end{tabular}
}
\end{sc}
\end{small}
\end{center}
\end{table}

\begin{figure}[t]
\begin{center}
\centerline{\includegraphics[width=\columnwidth]{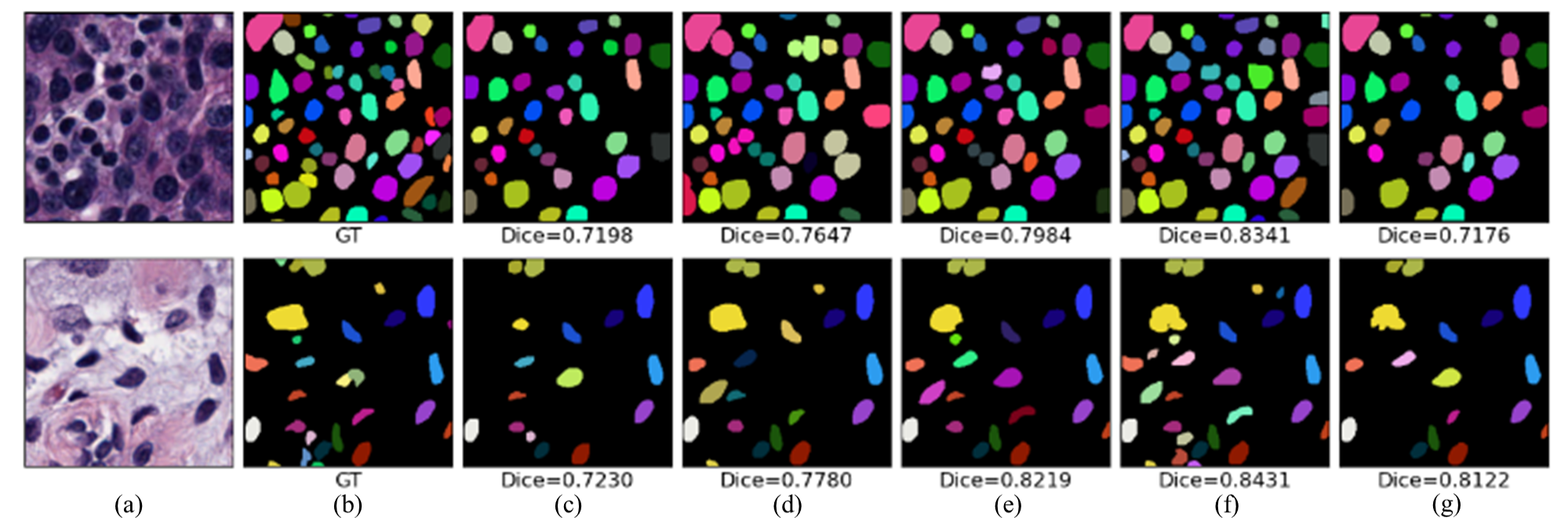}}
\caption{Outputs from different cycles on MONu. From left to right: the original images, the ground truth, the predictions of the best student from the $1^{\text{st}}$ cycle to the $5^{\text{th}}$ cycle.}
\label{fig:cycle}
\end{center}
\end{figure}

The core of cyclic learning is to build a bilateral knowledge sharing bridge between the front-end classification and the back-end nuclei instance segmentation, which establishes an MTL setting. We conduct extra experiments of unilaterally sharing weights from the back-end to the front-end at each cycle to demonstrate this point. Additionally, we evaluate the front-end classification performance with the F1-score to verify that bilateral knowledge sharing improves the performance of both tasks. The comparison of the 2 knowledge sharing methods (unilateral and bilateral) is shown in \autoref{tab:uni bi}, where the best results of the unilateral knowledge sharing mode are underlined. It is noted that the first cycle of the unilateral mode substantially shows the results of straightforwardly combining a CNN interpreting method and a semi-supervised learning method only with pseudo seeds. We can observe that the performance of straightforward combination ($1^{st}$ row) is patchy. The straightforward combination achieves scores of 0.7174, 7.1655, 0.5635, and 0.6571 for the Dice, HD, AJI, and Dice$_{obj}$, respectively. In contrast, our MTL-based cyclic learning achieves significantly better results with scores of 0.7739, 5.6703, 0.6357, and 0.7171 ($9^{th}$ row, Table 4), representing relative improvements of 7.88\%, 20.87\%, 12.81\%, and 9.13\%, respectively. Even the front-end F1-score is improved from 0.9294 to 0.9540, representing a relative improvement of 2.65\%. When continuing knowledge sharing, the best performance of the unilateral mode is inferior to that of the bilateral mode in both the front-end and the back-end. \autoref{fig:cycle} demonstrates that cyclic learning indeed gradually polishes the predicted masks. It makes a better matching from the predicted masks to the ground truths and yields better nuclei boundaries. The results in \autoref{tab:uni bi} demonstrate that the front-end and the back-end improve both with cyclic learning. But still, we notice that overtraining the system with cyclic learning may lead to overfitting to some extent ($5^{\text{th}}$ cycle in \autoref{tab:cycle-student} and \autoref{tab:uni bi}, last column in \autoref{fig:cycle}). 

\subsubsection{Front-end} 
A naive perception is that the closer the shape of pseudo seeds to the ground-truth masks generated by the front-end, the better the back-end semi-supervised instance segmentation would perform. But the fact is that the back-end performances are similar no matter where the generated pseudo seeds are good or bad. This point can be demonstrated by a controlled experiment, which is listed in \autoref{tab:front-end}. It shows that the performance of using LRP-generated pseudo seeds for the back-end is nearly equal to using the ground truths. 

\begin{figure}[t]
\begin{center}
\centerline{\includegraphics[width=\columnwidth]{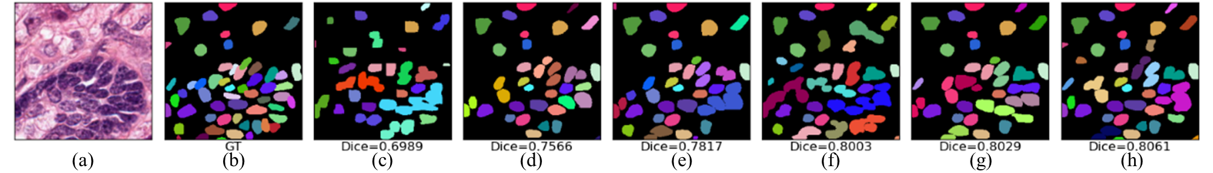}}
\caption{Outputs of cyclic leaning using different $K$. \textcolor{darkcyan}{(a)} original images; \textcolor{darkcyan}{(b)} the ground truth; \textcolor{darkcyan}{(c)} $K=3$; \textcolor{darkcyan}{(d)} $K=5$; \textcolor{darkcyan}{(e)} $K=10$; \textcolor{darkcyan}{(f)} $K=15$; \textcolor{darkcyan}{(g)} $K=20$; \textcolor{darkcyan}{(h)} $K=40$. }
\label{fig:seedK}
\end{center}
\end{figure}

To explore the influence of the total number of pseudo seeds $K$, we set $K$ as 3, 5, 10, 15, 20, and 40, respectively. The results are illustrated in \autoref{fig:seedK}. The Dice results are 0.6989, 0.7566, 0.7817, 0.8003, 0.8029, 0.8061 as $K$ increases from 3 to 40. It turns out that a larger $K$ does bring a positive influence to the back-end semi-supervised learning. However, the performance of $K=20$  is very close to the one of $K=40$, indicating $K$ is neither the crucial key to the performance gain over the nuclei instance segmentation.

\subsubsection{Back-end}

We conduct experiments on how vital it is to adopt a semi-supervised learning method as the back-end in our method. First, without cyclic learning, we adopt an ordinary Mask-RCNN for the back-end as the baseline, which is the first teacher $f_1^{S}$ without semi-supervised learning in cycle 1 presented in \autoref{tab:cycle-student}. In the same way, if we only add a semi-supervised strategy in the back-end, the back-end will quickly overfit, with the best Dice achieving merely 0.7181 ($f_2^{S}$ in cycle 1 in \autoref{tab:cycle-student}). But what if we only add cyclic learning into the baseline, i.e. conduct bilateral knowledge sharing between the front-end CNN and the back-end Mask-RCNN? The results of this experiment are shown in \autoref{tab:back-end}. It shows that even if you only add cyclic learning, the instance segmentation performance is already on par with only adding semi-supervised learning, which indicates that cyclic learning indeed learns a better backbone through circularly sharing knowledge. When we adopt semi-supervised methods as the back-end and add the cyclic learning process, the system is vastly improved, even close to the fully-supervised counterpart. To summarize, the back-end semi-supervised part is crucial for our approach to work, but it is the core of cyclic learning to unlock the potential of our approach.

 \subsection{Flexibility and Generality Analysis}

Cyclic learning is an MTL technique with flexibility and generality. In this section, we conduct experiments to validate the effectiveness of cyclic learning under different conditions.

\subsubsection{Different segmentation architectures}

Cyclic learning is a paradigm that can incorporate other advanced segmentation architectures as the back-end $f^S$, as long as these architectures possess general feature extracting backbones of applicability for the front-end classification. This characteristic is essential for maintaining the consistency of backbones at both ends. We conduct experiments on changing $f^S$ from Mask-RCNN to either Hover-net \cite{graham2019hover} or KG Instance \cite{yi2019multi} for cyclic learning and compare the results with their fully-supervised counterparts. For these experiments, the feature extracting backbone in the front-end $f^C$ is kept identical to that of $f^S$ for cyclic learning. The quantitative results are summarized in \autoref{tab:diff fs}, with the first and third rows directly adopted from \cite{graham2019hover}, while the second and fourth rows present our implementation results. We use the default setting when training the fully-supervised Hover-net and KG Instance.

\begin{figure}[t]
\begin{center}
\centerline{\includegraphics[width=\columnwidth]{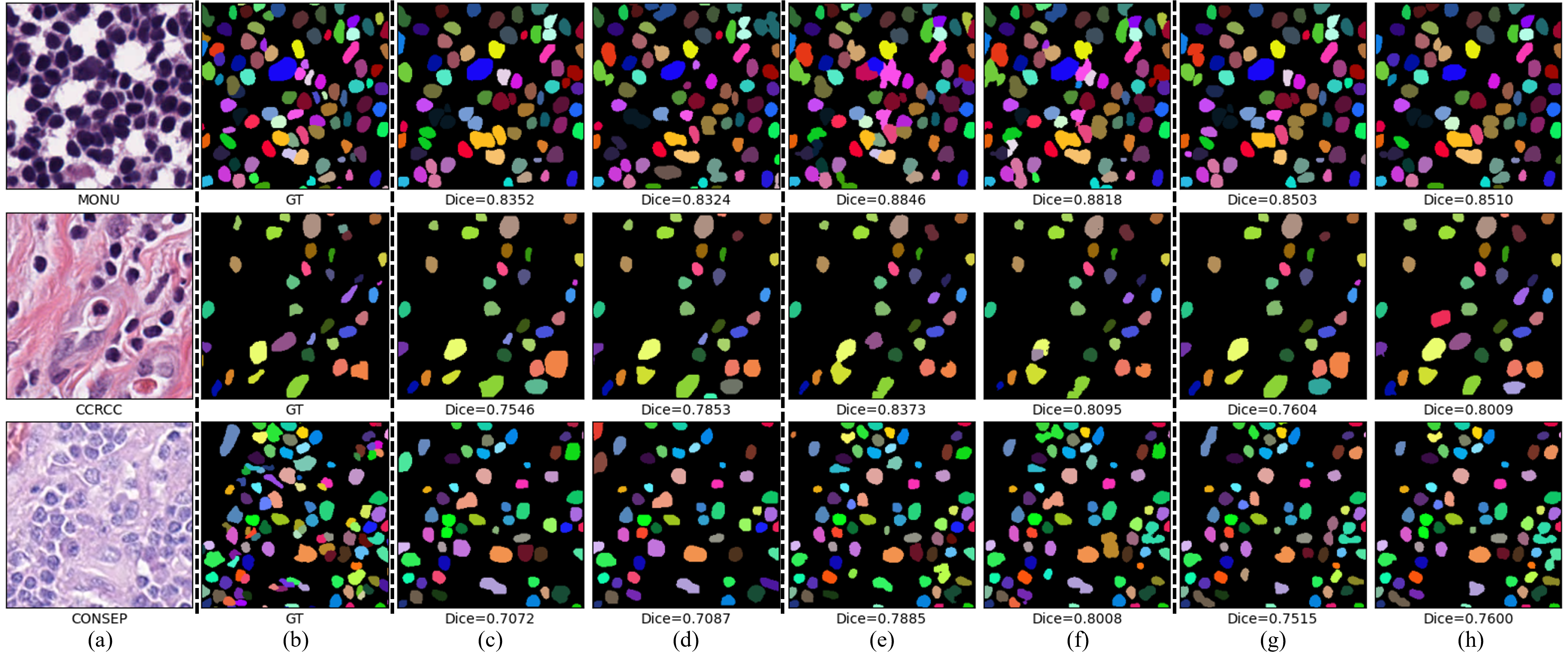}}
\caption{Sample outputs of cyclic learning (CL) using different architectures as the back-end $f^S$ and the outputs of the corresponding fully-supervised counterparts. From top to bottom: MONu, CCRCC, CoNSeP. \textcolor{darkcyan}{(a)} original images; \textcolor{darkcyan}{(b)} the ground truth; \textcolor{darkcyan}{(c)} CL-Mask-RCNN; \textcolor{darkcyan}{(d)} fully-supervised Mask-RCNN; \textcolor{darkcyan}{(e)} CL-Hover-net; \textcolor{darkcyan}{(f)} fully-supervised Hover-net; \textcolor{darkcyan}{(g)} CL-KG Instance; \textcolor{darkcyan}{(h)} fully-supervised KG Instance. }
\label{fig:difffs}
\end{center}
\end{figure}

As shown in \autoref{tab:diff fs}, the cyclic learning using different architectures as $f^S$ performs close to the fully-supervised counterpart as well, and sometimes even achieves better performance on certain metrics. Corresponding sample outputs are illustrated in \autoref{fig:difffs}. The figure indicates that fully-supervised Hover-net and KG Instance have better discrimination ability than Mask-RCNN, as they are specially devised for nuclei segmentation. The numerical results and visualizations demonstrate that cyclic learning is adaptable to different segmentation architectures.

\subsubsection{Transformer-based backbone}

In \autoref{tab:backbone}, cyclic learning has already demonstrated its generality with CNN backbone architecture. At present, many advanced methods have adopted the transformer \cite{vaswani2017attention} as the backbone, which has achieved excellent performance. To validate the compatibility of cyclic learning with transformer-based backbone, we adopt the swin-transformer \cite{liu2021swin} as the backbone of Mask-RCNN under fully-supervised mode and cyclic learning. Without loss of generality, we use swin-tiny for validation and initialize the backbone with \textit{``swin\_tiny\_patch4\_window7\_224''}. Following Liu \textit{et al.} \cite{liu2021swin}, we feed the hierarchical features of swin-transformer into the FPN of Mask-RCNN for network construction. We use attention rollout \cite{abnar2020quantifying} as the interpreting method for the front-end. The results are listed in \autoref{tab:transformer}, where the quantitative results of cyclic learning are on the verge of fully-supervised counterparts. These results indicate that cyclic learning is still highly adaptable when the backbone is transformer-based.

\subsubsection{Synthetic data as auxiliary training data}

To achieve better adaptation performance in nuclei segmentation, some methods propose using synthetic data as extra data \cite{kromp2021evaluation, mahmood2019deep}. It is wondered whether this kind of method can also enhance cyclic learning. Mahmood \textit{et al.} propose deep adversarial training (DAT), which utilizes an unpaired generative adversarial network to transfer artificial polygon masks to synthetic H\&E images, thereby creating auxiliary data for fully-supervised training \cite{mahmood2019deep}. We employ DAT to generate synthetic data with masks and add them to the training set of MONu. We conduct a comparison between fully-supervised learning and cyclic learning, with and without the synthetic data generated by DAT, using Mask-RCNN as the segmentation network. The results are presented in \autoref{tab:synthetic data}.

As reported by Mahnood \textit{et al.} \cite{mahmood2019deep}, DAT helps to improve the fully-supervised method. Surprisingly, even though only the image-level labels of synthetic data are provided, DAT contributes to cyclic learning as well. The result further highlights the generality of cyclic learning.

\section{Conclusion}
In this work, we propose a novel weakly supervised method to solve dense nuclei instance segmentation on histopathology images under image-level supervision. Our method utilizes cyclic learning based on MTL to solve this task. Cyclic learning casts this task as a front-end classification task and a back-end semi-supervised task. It circularly shares knowledge between the front-end classifier and the back-end semi-supervised part, which allows the whole system to fully extract the underlying information from image-level labels and converge to a better optimum. Experiments on MONu, CCRCC, and CoNSeP show that our image-level weakly supervised method is close to the fully-supervised counterpart in performance. In addition, cyclic learning is adaptable to different feature extracting backbones, including CNN and transformer, and is compatible with different segmentation architectures, such as Mask-RCNN, Hover-net, and KG Instance. Future works will focus on applying the algorithm to other dense instance segmentation in microscopic images.


\normalem
\bibliographystyle{IEEEtran}
\bibliography{tmi.bib}
\end{document}